\RequirePackage{snapshot}
\documentclass[10pt,twocolumn,letterpaper]{article}

\usepackage{microtype}
\usepackage{graphicx}
\usepackage{csquotes}
\usepackage{subcaption}
\usepackage{booktabs} %
\usepackage{amsmath}
\usepackage{duckuments}
\usepackage{mathtools}
\usepackage{makecell}
\usepackage{multirow}
\usepackage{comment}
\usepackage{graphbox}
\usepackage{tabularx}
\usepackage{float}
\usepackage{amsfonts}
\usepackage{placeins}
\usepackage[table]{xcolor}
\usepackage[export]{adjustbox} %
\usepackage{pgfplots,pgfplotstable}
\usepgfplotslibrary{fillbetween}
\pgfplotsset{compat=1.18} 

\usepackage[pagenumbers]{cvpr} %
\pgfplotsset{xtick style={draw=none}}
\pgfplotsset{ytick style={draw=none}}
\pgfplotsset{major grid style={gray!40}}

\definecolor{C0}{rgb}{0.121569, 0.466667, 0.705882}
\definecolor{C1}{rgb}{1.000000, 0.498039, 0.054902}
\definecolor{C2}{rgb}{0.172549, 0.627451, 0.172549}
\definecolor{C3}{rgb}{0.839216, 0.152941, 0.156863}
\definecolor{C4}{rgb}{0.580392, 0.403922, 0.741176}
\definecolor{C5}{rgb}{0.549020, 0.337255, 0.294118}
\definecolor{C6}{rgb}{0.890196, 0.466667, 0.760784}
\definecolor{C7}{rgb}{0.498039, 0.498039, 0.498039}
\definecolor{C8}{rgb}{0.737255, 0.741176, 0.133333}
\definecolor{C9}{rgb}{0.090196, 0.745098, 0.811765}
\definecolor{C10}{rgb}{0.5, 0, 0.13} %
\definecolor{C11}{rgb}{0.9, 0, 0.13}

\definecolor{B0}{rgb}{0.121569, 0.466667, 0.705882}
\definecolor{B1}{rgb}{0.221569, 0.566667, 0.805882}
\definecolor{B2}{rgb}{0.321569, 0.666667, 0.905882}
\definecolor{B3}{rgb}{0.421569, 0.766667, 1.000000}
\definecolor{B4}{rgb}{0.521569, 0.866667, 1.000000}
\definecolor{B5}{rgb}{0.621569, 0.966667, 1.000000}

\newcommand{\teaser}{
\includegraphics[width=\textwidth]{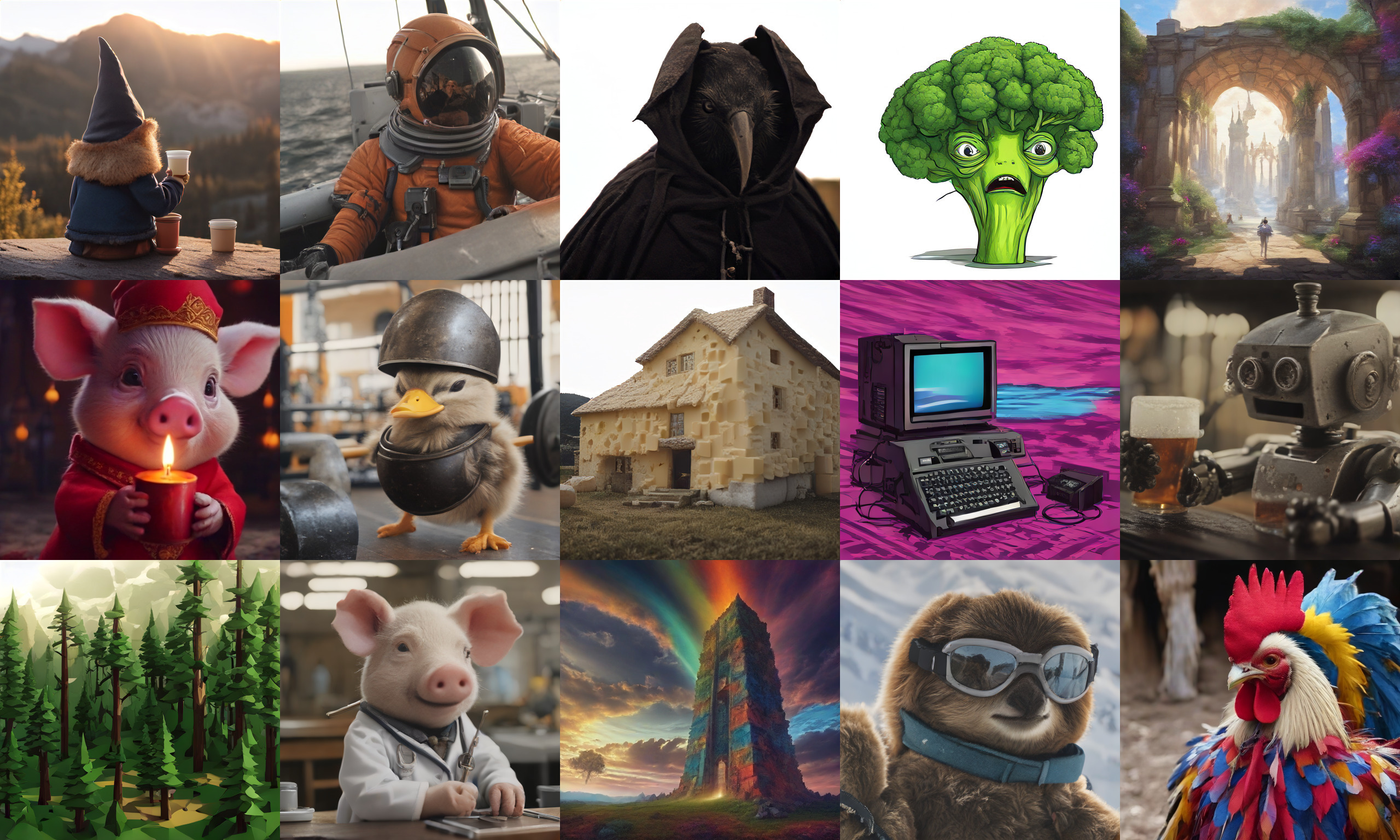}
\captionof{figure}{
\textbf{Generating high-fidelity $512^2$ images in a single step.}
All samples are generated with a single U-Net evaluation trained with adversarial diffusion distillation (ADD).
}
\label{fig:teaser}
\vspace{1.5em} 
}

\newcommand{\qualitativecomp}{
  \begin{figure*}[htbp]
    \centering
    \small
\begin{tabular}{lll}
    &
    \parbox[b]{.45\linewidth}{\centering \emph{A cinematic shot of a professor sloth wearing a tuxedo at a BBQ party.}}
    \vspace{0.5em}
    &
    \parbox[b]{.45\linewidth}{\centering \emph{A high-quality photo of a confused bear in calculus class. The bear is wearing a party hat and steampunk armor.}}
    \\
  \rotatebox[origin=c]{90}{\makecell{\modelshort-XL \\ (1 step)}} &
  \includegraphics[width=.45\linewidth,valign=m]{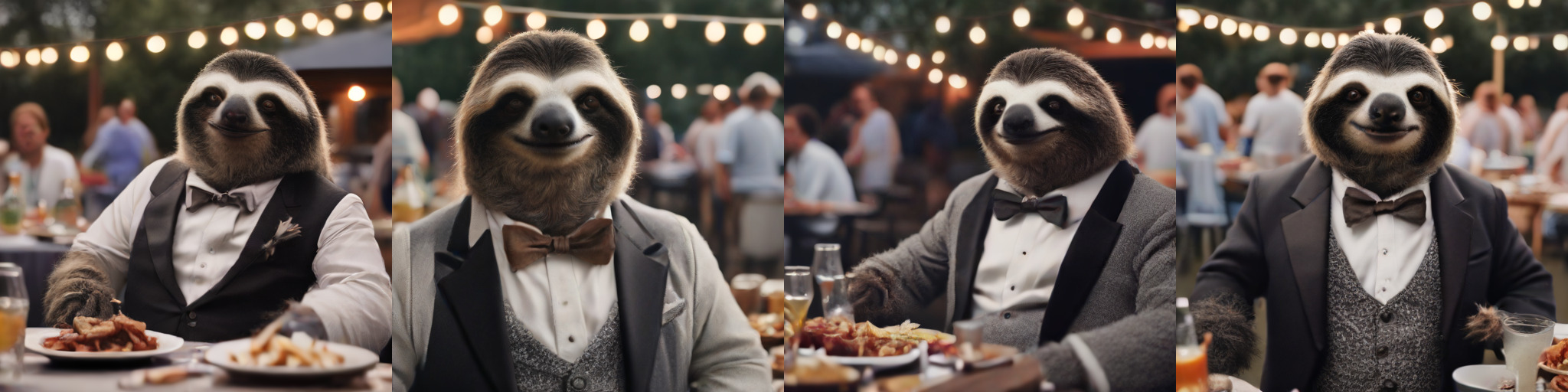}
  &
  \includegraphics[width=.45\linewidth,valign=m]{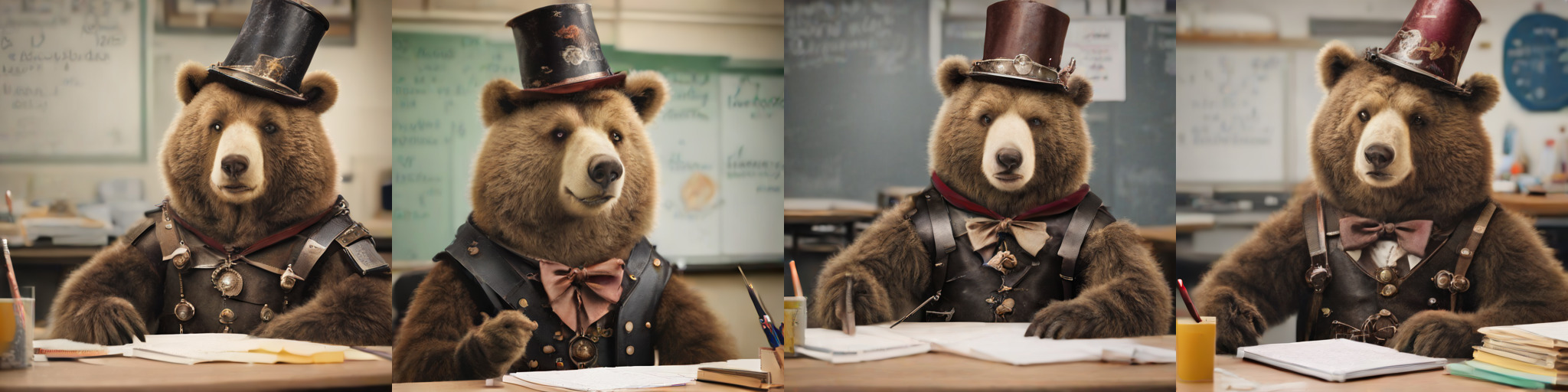}
  \\
  \rotatebox[origin=c]{90}{\makecell{LCM-XL \\ (1 step)}} &
  \includegraphics[width=.45\linewidth,valign=m]{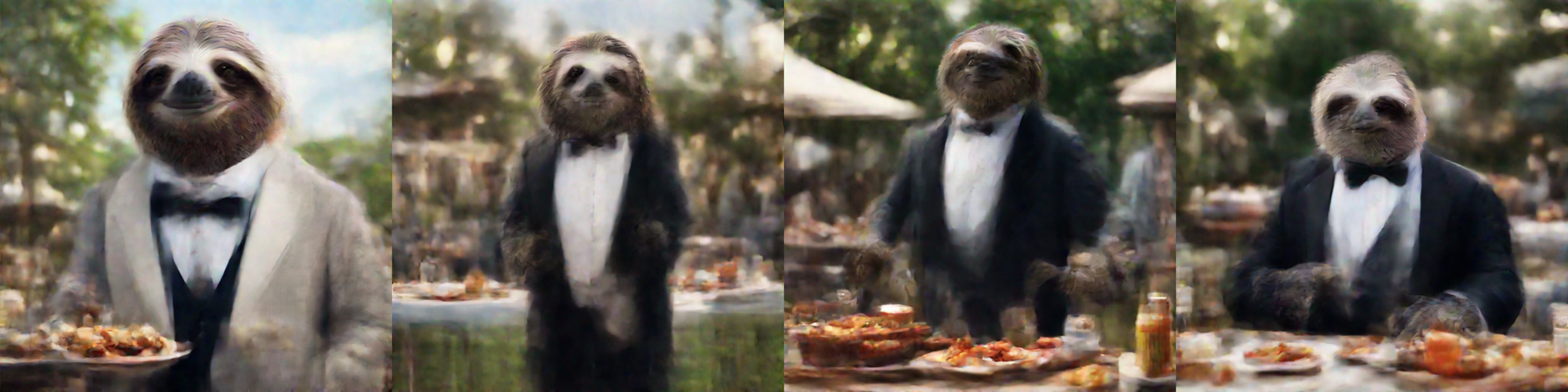}
  &
  \includegraphics[width=.45\linewidth,valign=m]{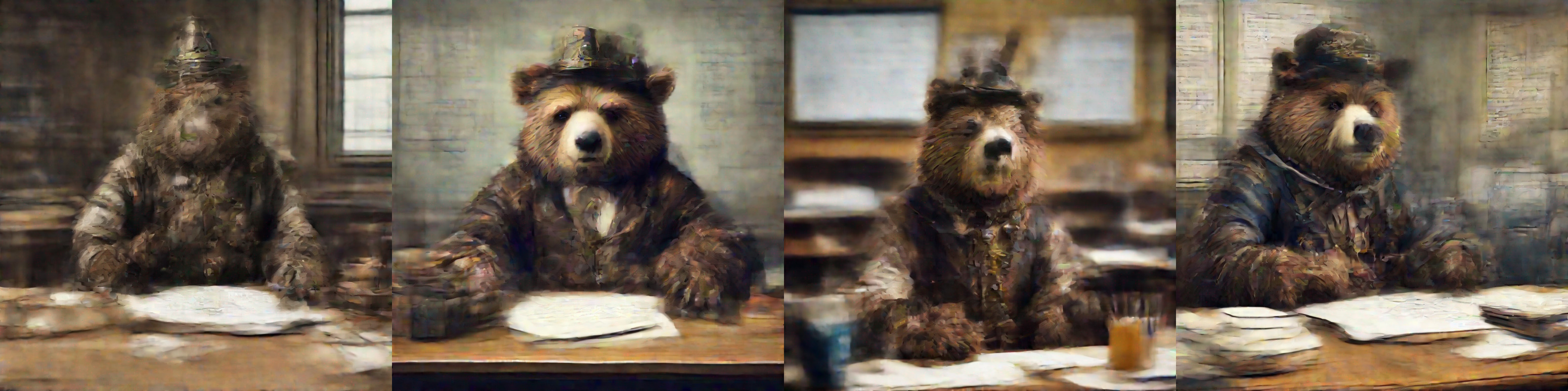}
  \\
  \rotatebox[origin=c]{90}{\makecell{LCM-XL \\ (2 steps)}} &
  \includegraphics[width=.45\linewidth,valign=m]{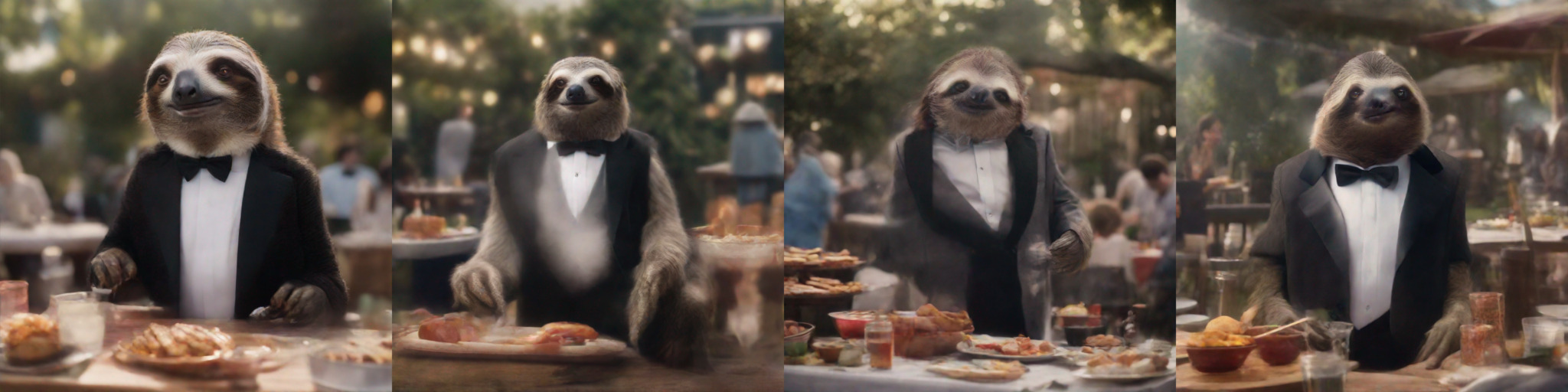}
  &
  \includegraphics[width=.45\linewidth,valign=m]{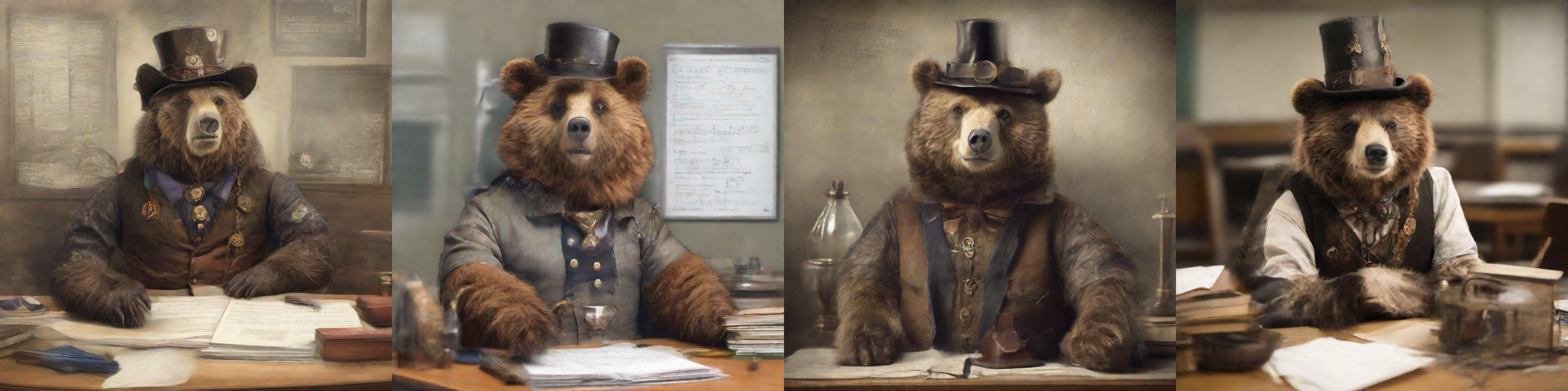}
  \\
  \rotatebox[origin=c]{90}{\makecell{LCM-XL \\ (4 steps)}} &
  \includegraphics[width=.45\linewidth,valign=m]{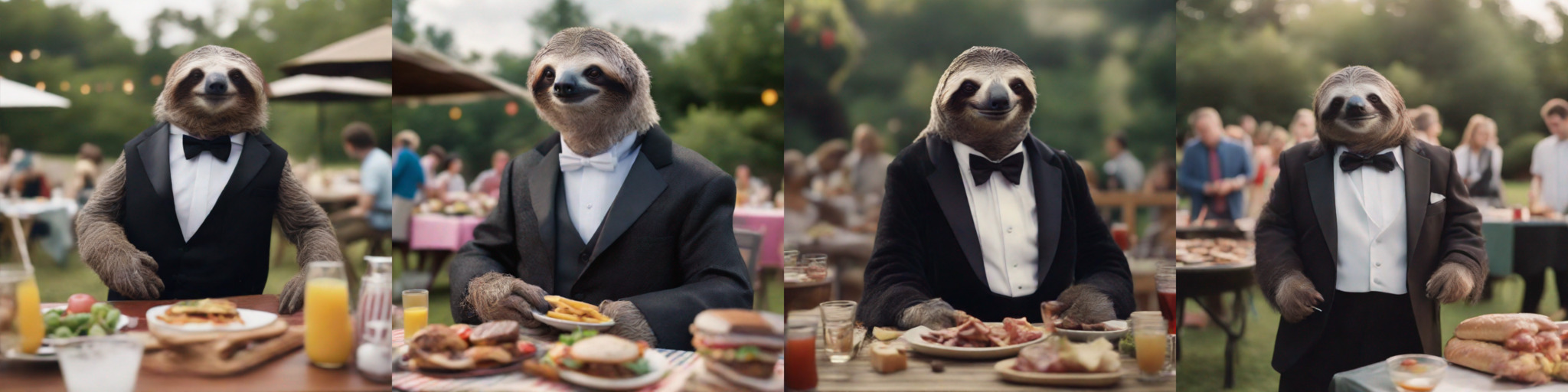}
  &
  \includegraphics[width=.45\linewidth,valign=m]{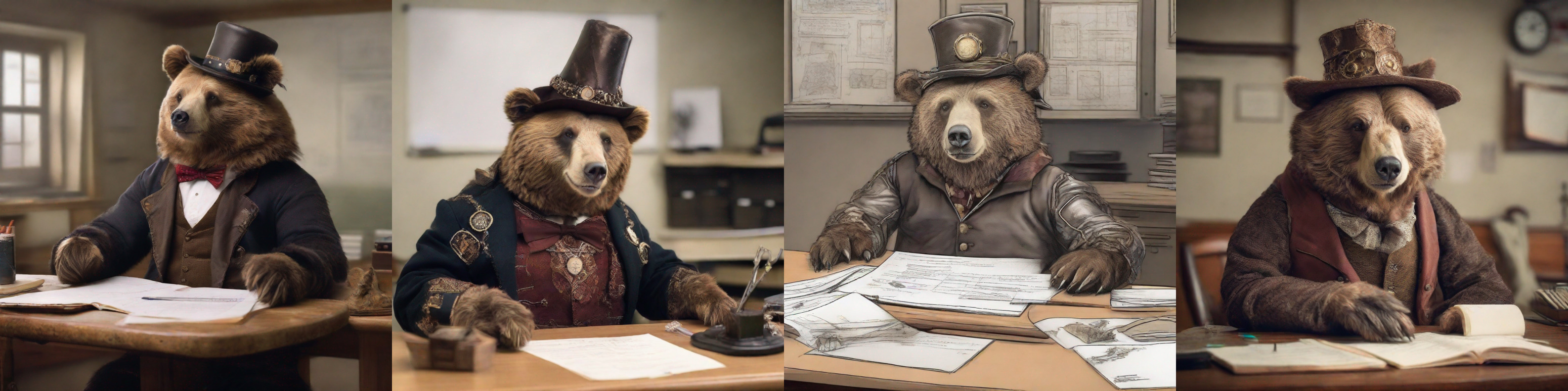}
  \\
  \rotatebox[origin=c]{90}{\makecell{StyleGAN-T++ \\ (1 step)}} &
  \includegraphics[width=.45\linewidth,valign=m]{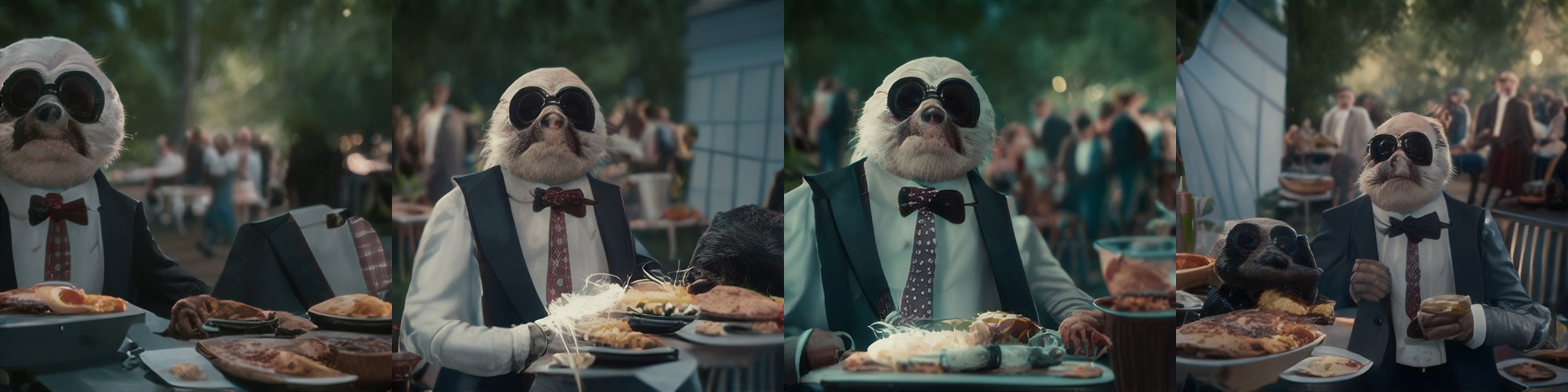}
  &
  \includegraphics[width=.45\linewidth,valign=m]{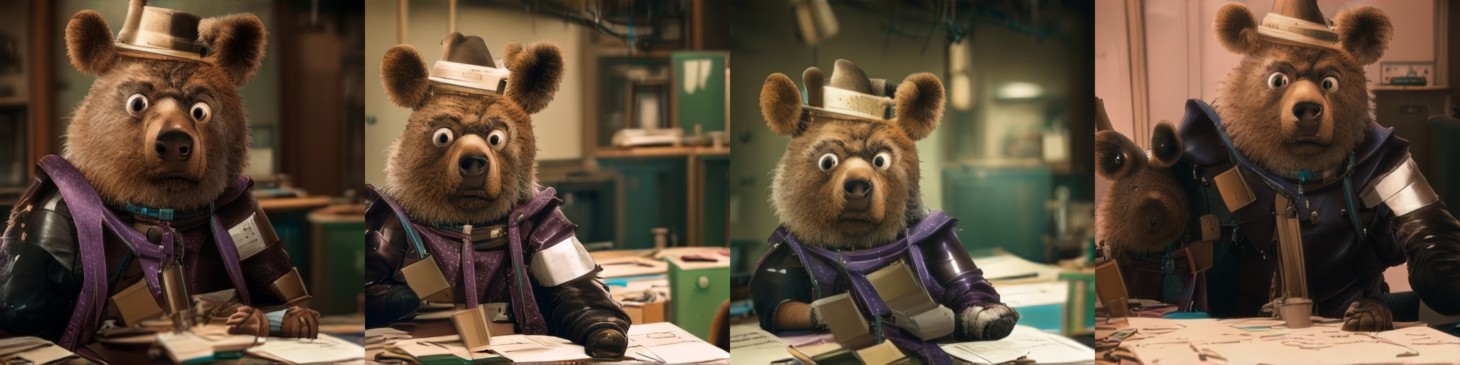}
  \\
  \rotatebox[origin=c]{90}{\makecell{InstaFlow\\ (1 step)}} &
  \includegraphics[width=.45\linewidth,valign=m]{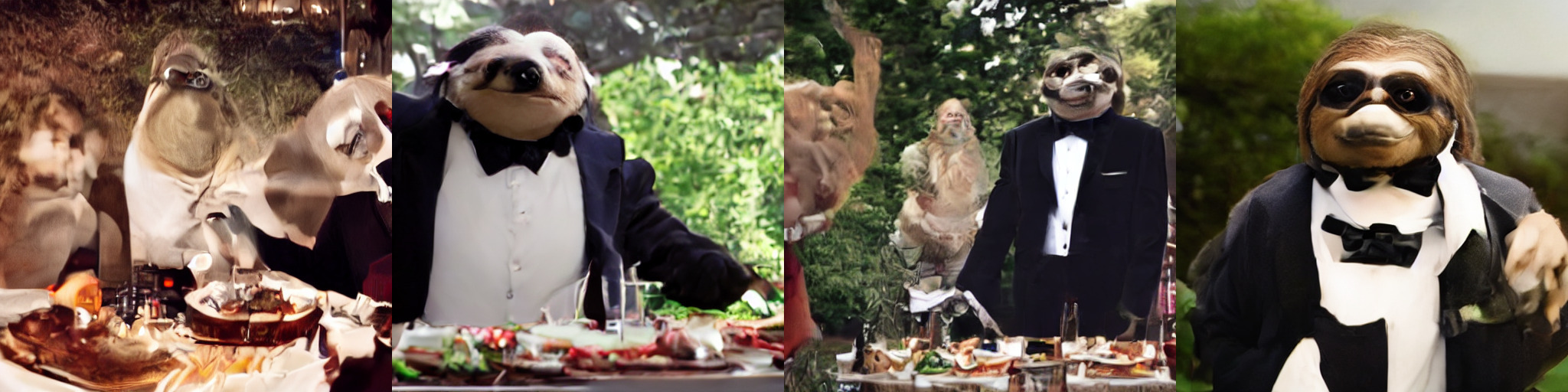}
  &
  \includegraphics[width=.45\linewidth,valign=m]{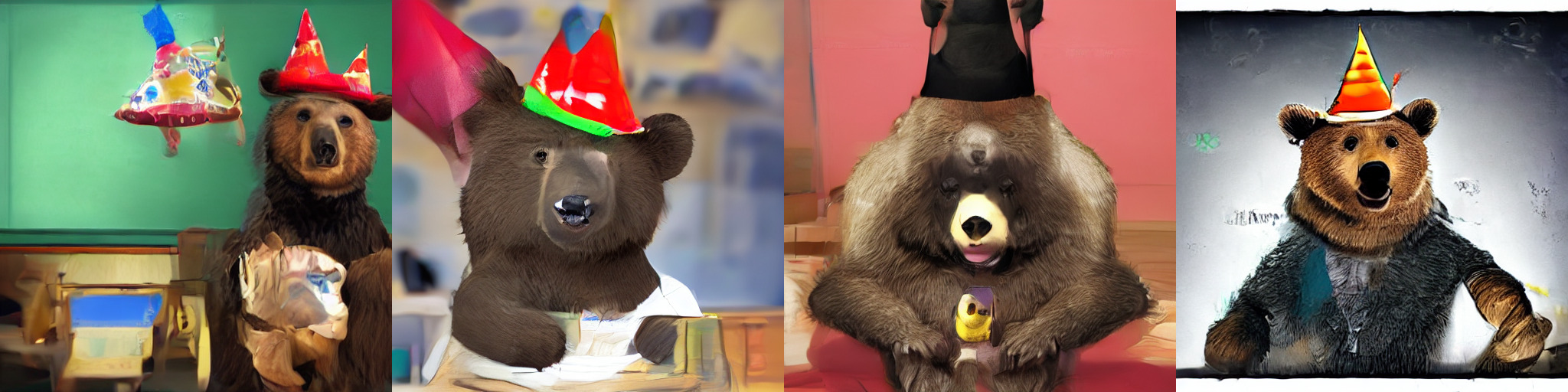}
  \\
    \rotatebox[origin=c]{90}{\makecell{OpenMUSE\\ (16 steps)}} &
  \includegraphics[width=.45\linewidth,valign=m]{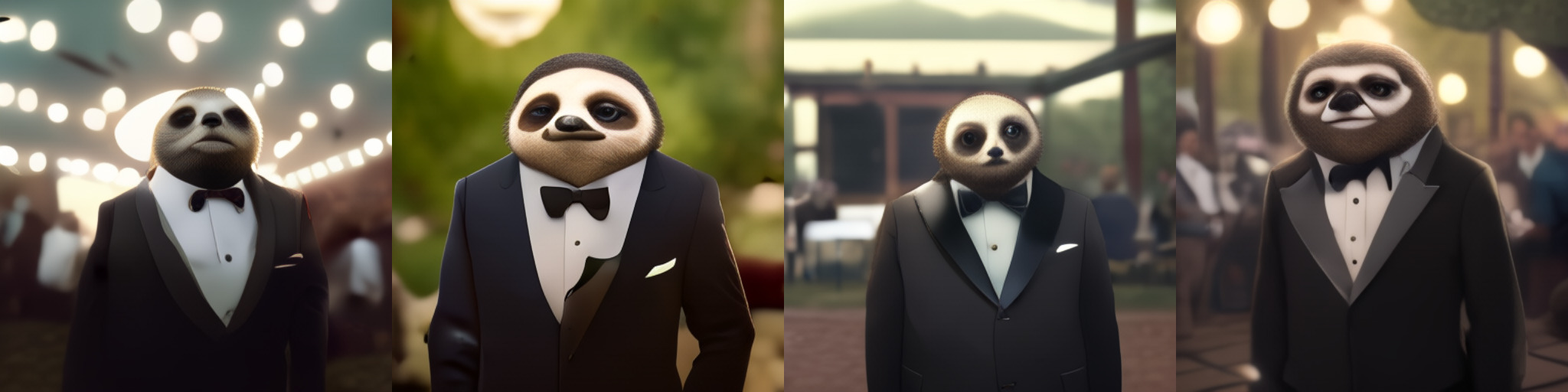}
  &
  \includegraphics[width=.45\linewidth,valign=m]{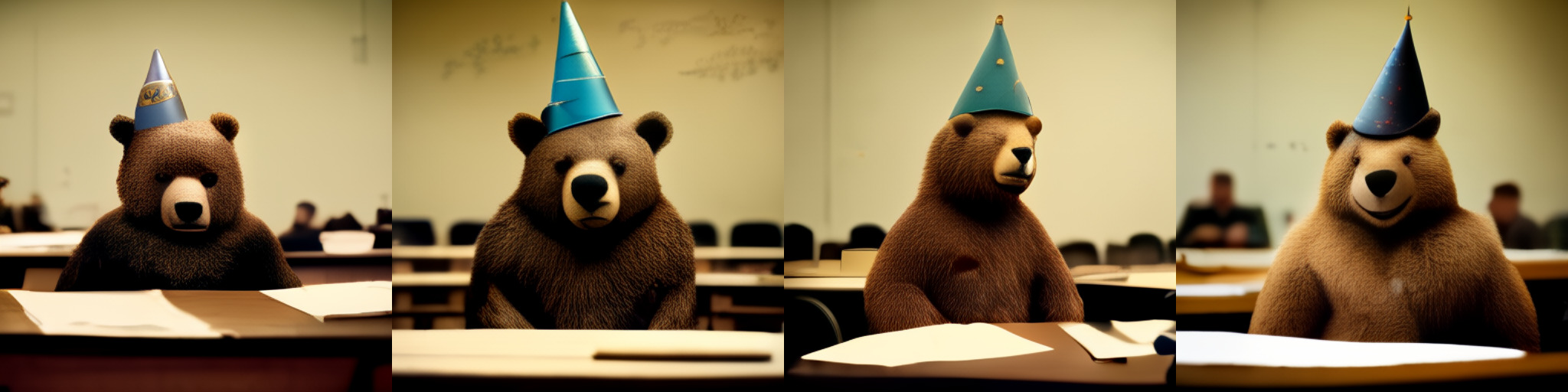}
  \\
\end{tabular}
\caption{
\textbf{Qualitative comparison to state-of-the-art fast samplers.}
Single step samples from our \modelshort-XL (top) and LCM-XL~\citep{Luo2023LCMLoRAAU}, our custom StyleGAN-T~\cite{sauer2023stylegan} baseline, InstaFlow~\citep{liu2023instaflow} and MUSE. For MUSE, we use the \emph{OpenMUSE} implementation and default inference settings with 16 sampling steps. For LCM-XL, we sample with 1, 2 and 4 steps. Our model outperforms all other few-step samplers in a single step.
\label{fig:qualitativecomp}
}
\end{figure*}
}

\newcommand{\qualitativeteacher}{
  \begin{figure*}[htbp]
    \centering
    \small
\begin{tabular}{lll}
&
\parbox[b]{.45\linewidth}{\centering \emph{A cinematic shot of a little pig priest wearing sunglasses.}}
\vspace{0.5em}
&
\parbox[b]{.45\linewidth}{\centering \emph{A photograph of the inside of a subway train. There are frogs sitting on the seats. One of them is reading a newspaper. The window shows the river in the background.}}
\\
\rotatebox[origin=c]{90}{\makecell{\modelshort-XL\\ (4 steps)}} &
\includegraphics[width=.45\linewidth,valign=m]{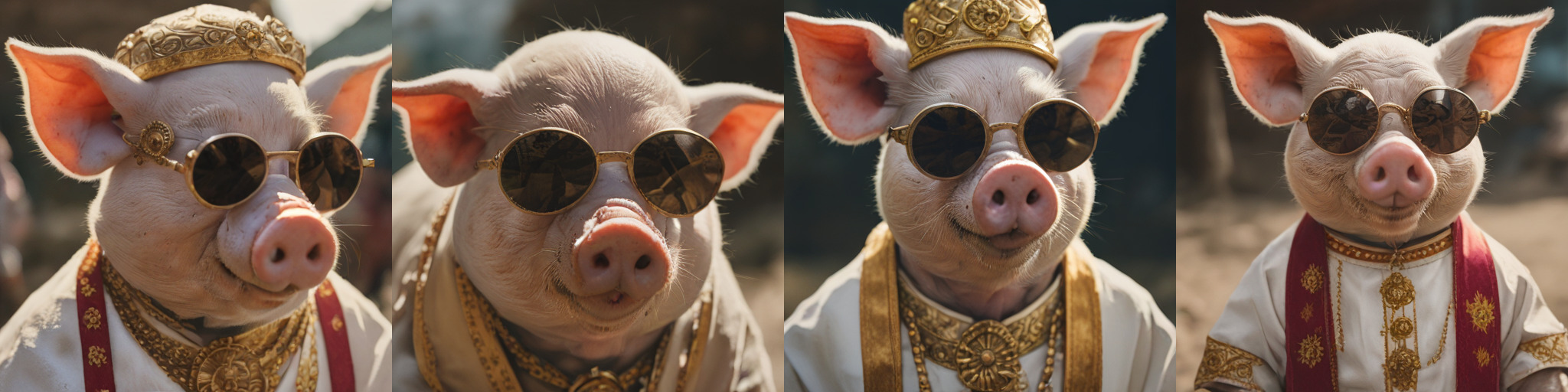}
&
\includegraphics[width=.45\linewidth,valign=m]{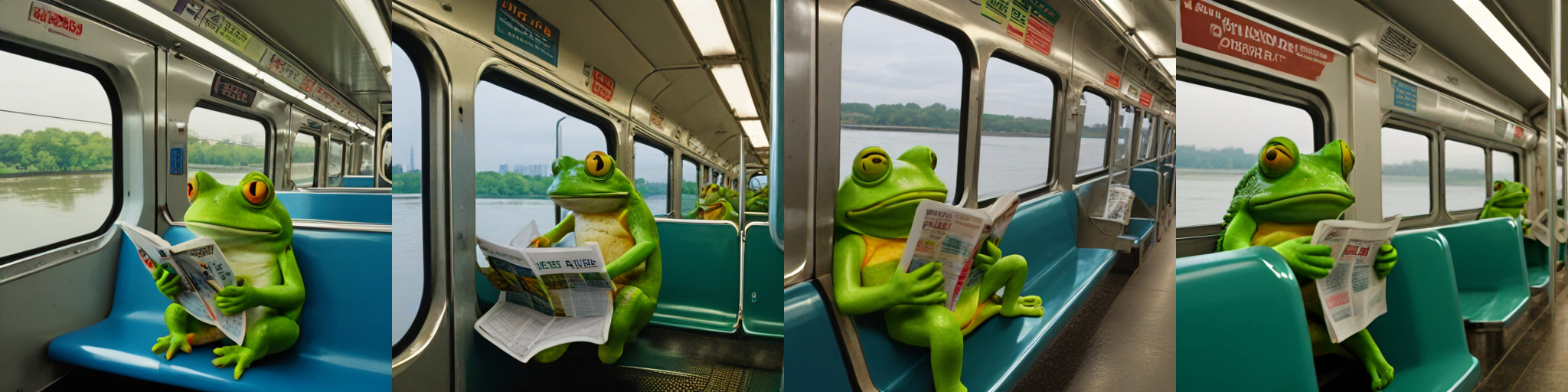}
\\
\rotatebox[origin=c]{90}{\makecell{SDXL-Base\\ (50 steps)}} &
\includegraphics[width=.45\linewidth,valign=m]{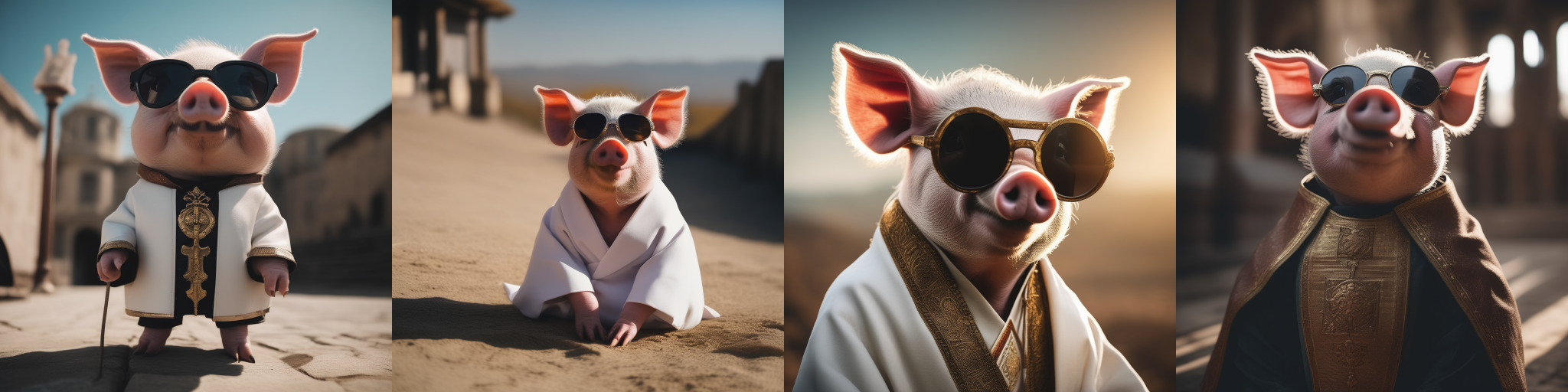}
&
\includegraphics[width=.45\linewidth,valign=m]{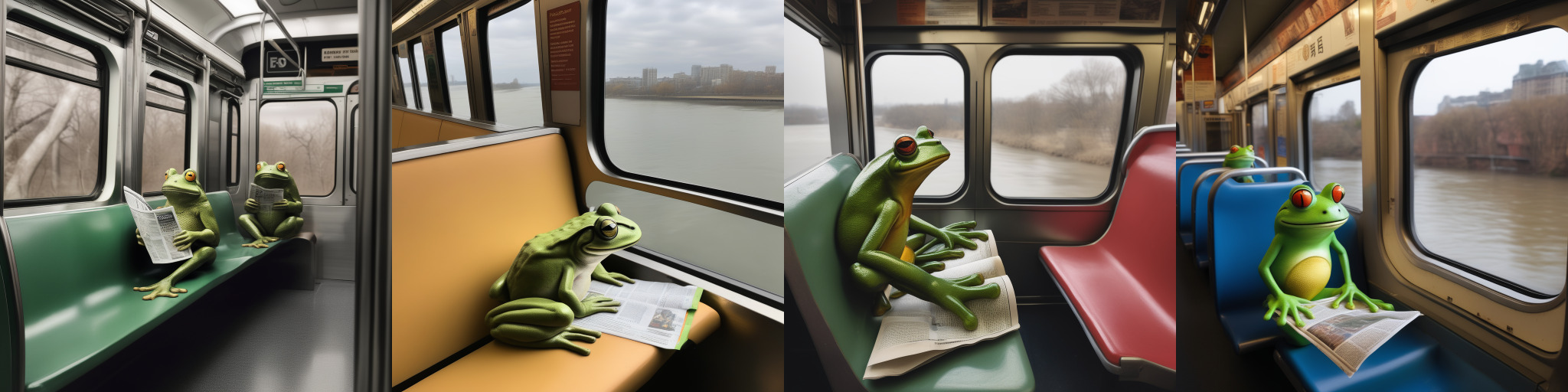}
\\
\addlinespace[10pt]
&
\parbox[b]{.45\linewidth}{\centering \emph{A photo of an astronaut riding a horse in the forest. There is a river in front of them with water lilies.}}
\vspace{0.5em}
&
\parbox[b]{.45\linewidth}{\centering \emph{A photo of a cute mouse wearing a crown.}}
\\
\rotatebox[origin=c]{90}{\makecell{\modelshort-XL\\ (4 steps)}} &
\includegraphics[width=.45\linewidth,valign=m]{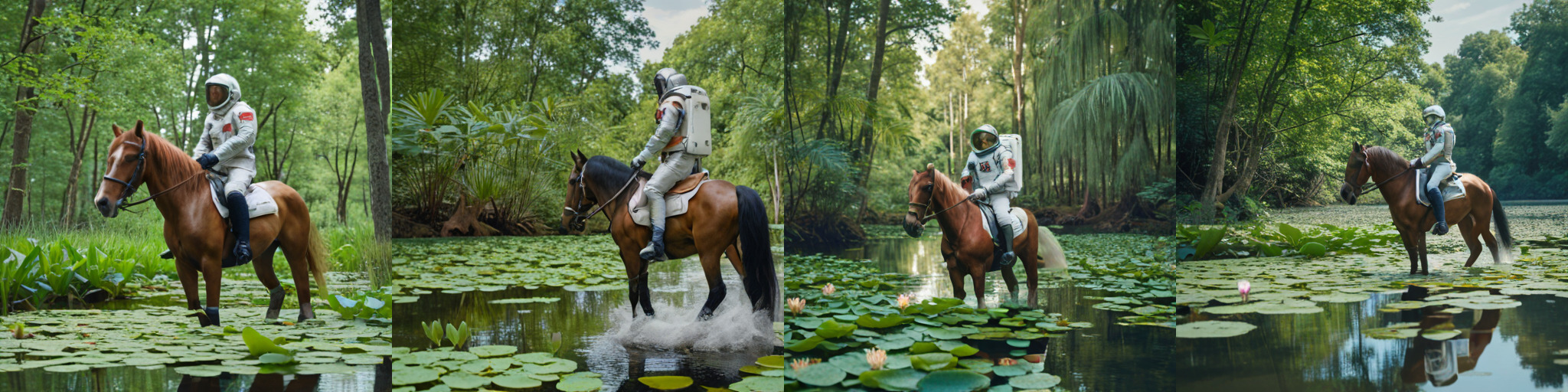}
&
\includegraphics[width=.45\linewidth,valign=m]{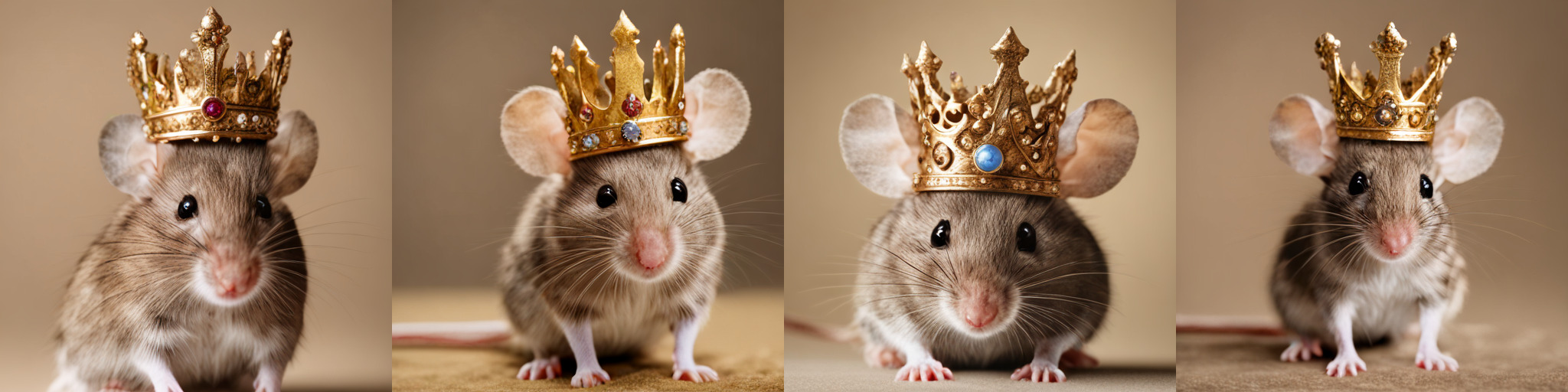}
\\
\rotatebox[origin=c]{90}{\makecell{SDXL-Base\\ (50 steps)}} &
\includegraphics[width=.45\linewidth,valign=m]{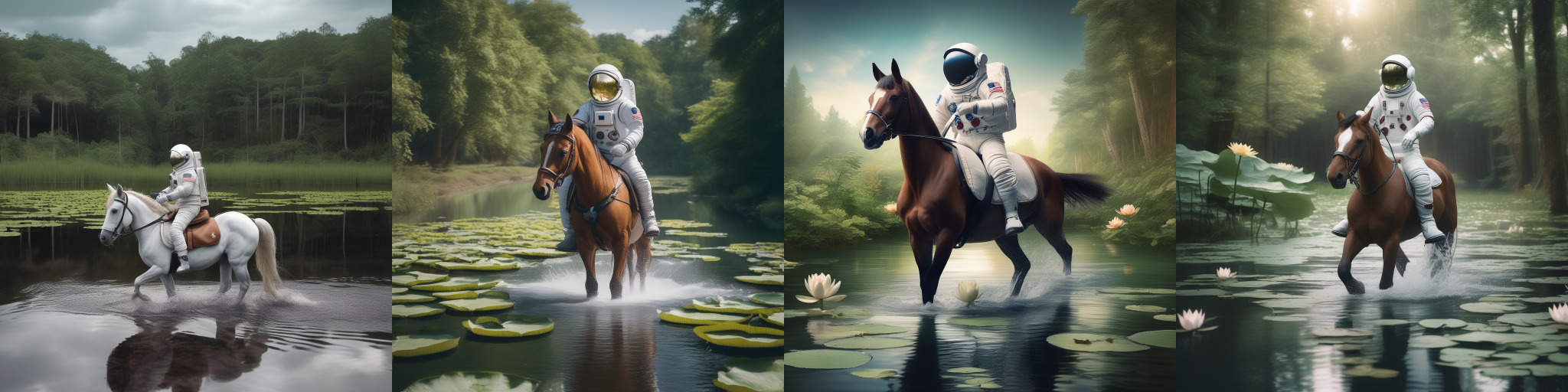}
&
\includegraphics[width=.45\linewidth,valign=m]{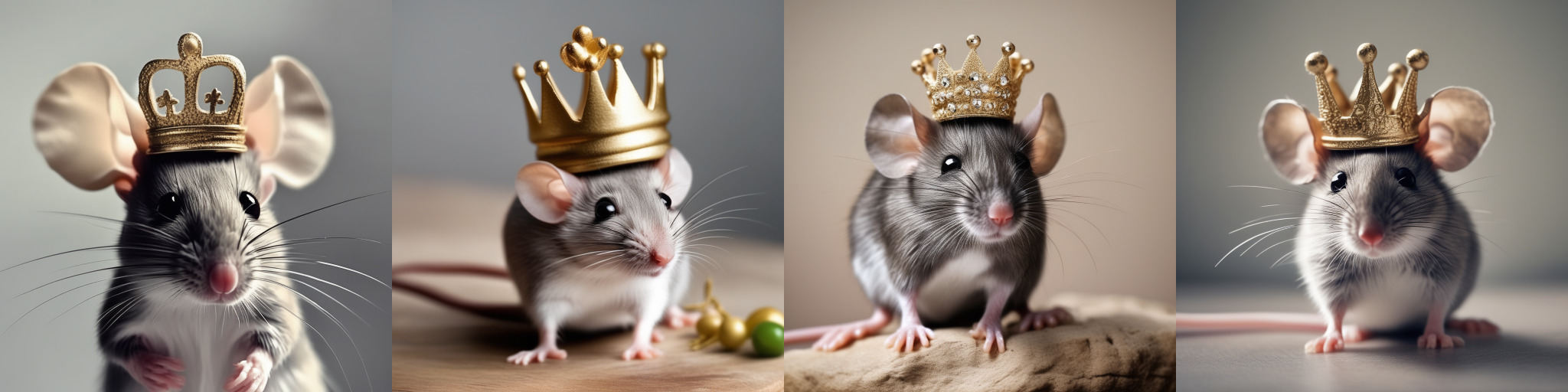}
\\
\end{tabular}
\caption{
\textbf{Qualitative comparison to the teacher model.}
ADD-XL can outperform its teacher model SDXL in the multi-step setting.
The adversarial loss boosts realism, particularly enhancing textures (fur, fabric, skin) while reducing oversmoothing, commonly observed in diffusion model samples.
ADD-XL's overall sample diversity tends to be lower.
\label{fig:qualitativeteacher}
}
\end{figure*}
}

\newcommand{\system}{
\begin{figure}[htbp]{
\includegraphics[trim={4.cm 15cm 2.3cm 0.5},clip,width=0.49\textwidth]{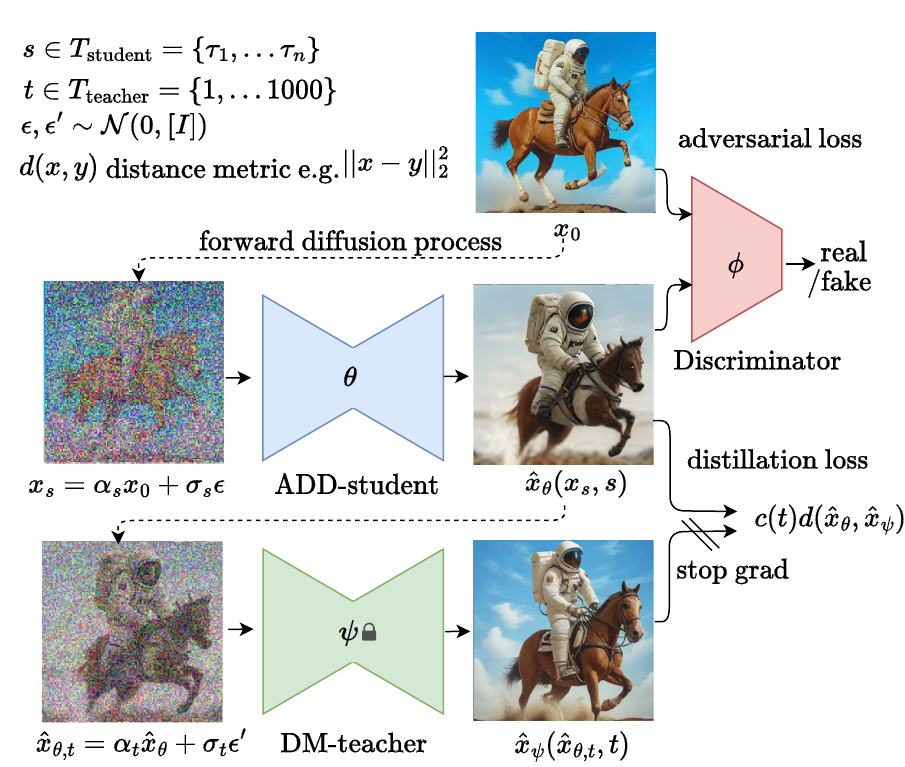}
}
\caption{
\textbf{Adversarial Diffusion Distillation.}
The ADD-student is trained as a denoiser that receives diffused input images $x_s$ and outputs samples $\hat{x}_{\theta}(x_s, s)$ and optimizes two objectives:
a) adversarial loss: the model aims to fool a discriminator which is trained to distinguish the generated samples $\hat{x}_{\theta}$ from real images $x_{0}$.
b) distillation loss: the model is trained to match the denoised targets $\hat{x}_{\psi}$ of a frozen DM teacher.
}
\label{fig:system}
\end{figure}
}

\newcommand{\teaserPoint}[5]{ %
  \addplot[#1, only marks, mark size=2pt, forget plot] coordinates {(#2,#3)};
  \node at (axis cs:#2,#3) [#1, anchor={#5}] {#4};
}

\newcommand{\eloplot}{
\begin{figure}[htbp]
\centering%
\resizebox{\linewidth}{!}{%
\begin{tikzpicture}%
\begin{axis}[
  width=100mm, height=66mm,
  xlabel={ELO $\uparrow$}, xmin={850}, xmax={1200}, xmode={linear}, 
  xtick={900, 950, 1000, 1050, 1100, 1150, 1200},
  ylabel={Inference speed [s] $\downarrow$}, ymin={0}, ymax={15}, y coord trafo/.code=\pgfmathparse{##1^0.5}, ytick={0, 1, 3, 5, 10, 12}, yticklabels={$0$, $1$, $3$, $5$, $10$, $12$},
  grid={major}, 
]

\teaserPoint{C0}{978}{10.49}{\makecell{IF-XL\\(150 steps)}}{west}
\teaserPoint{C0}{873}{0.849}{\makecell{OpenMUSE\\(16 steps)}}{south west}
\teaserPoint{C0}{981}{0.408}{\makecell{LCM-XL\\(4 steps)}}{south west}
\teaserPoint{C0}{1051}{3.557}{\makecell{SDXL\\(50 steps)}}{south west}
\teaserPoint{C11}{1009}{0.108}{\makecell{ADD-XL\\(1 step)}}{west}
\teaserPoint{C11}{1107}{0.408}{\makecell{ADD-XL\\(4 steps)}}{south west}

\end{axis}
\end{tikzpicture}%
}%
\vspace*{-2mm}%
\caption{%
\textbf{Performance vs. speed.} 
We visualize the results reported in~\figref{fig:humanevalmultiple} in combination with the inference speeds of the respective models.
The speeds are calculated for generating a single sample at resolution 512x512 on an A100 in mixed precision.
}%
\label{fig:eloplot}
\end{figure}
}

\newcommand{\humanevalallsingle}{
\begin{figure*}[t]
{
\centering
\begin{minipage}{0.49\linewidth}{
\begin{center}
\includegraphics[width=\linewidth]{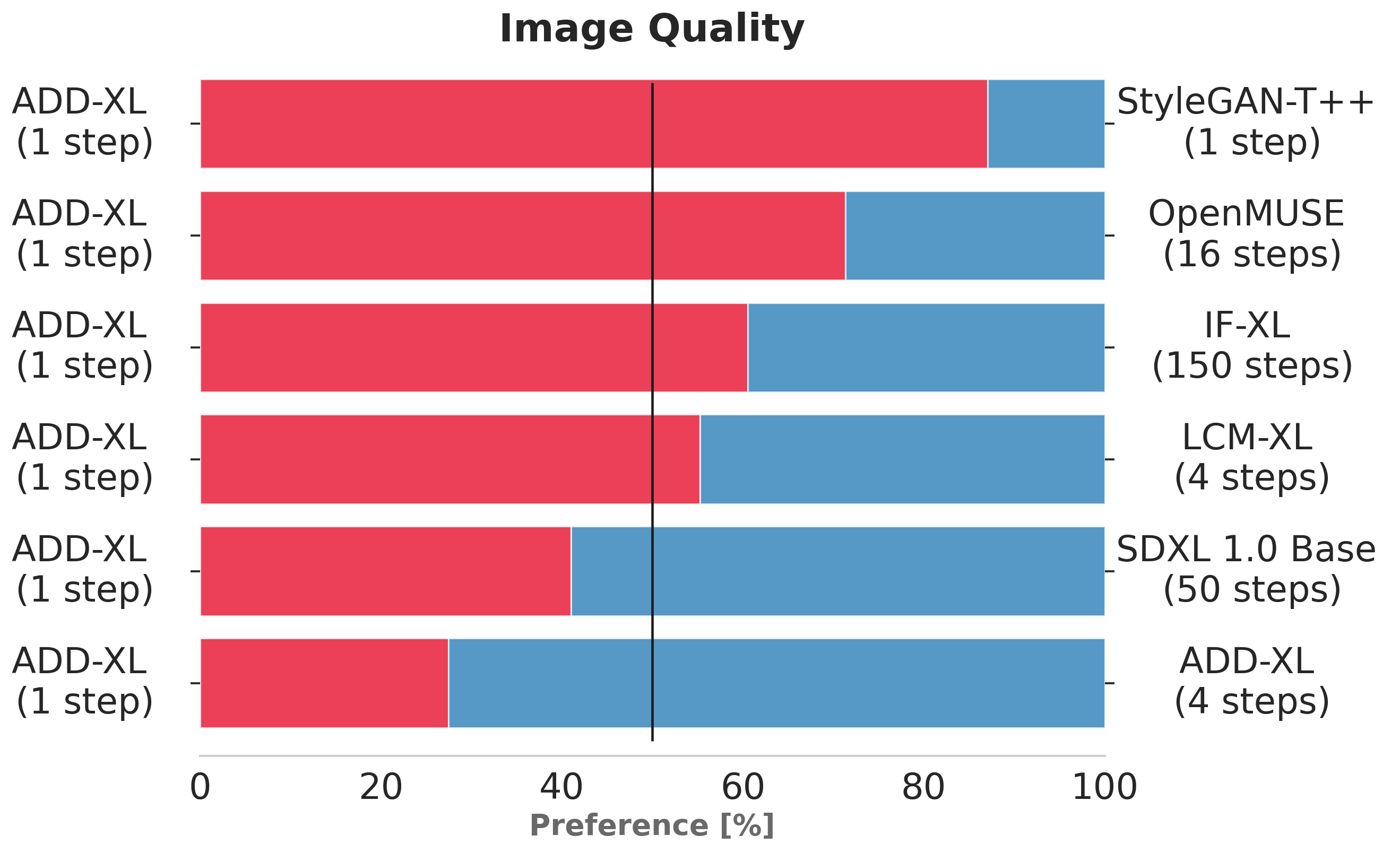}
\end{center}}\end{minipage}
}
\hfill
{
\begin{minipage}{0.49\linewidth}{
\begin{center}
\includegraphics[width=\linewidth]{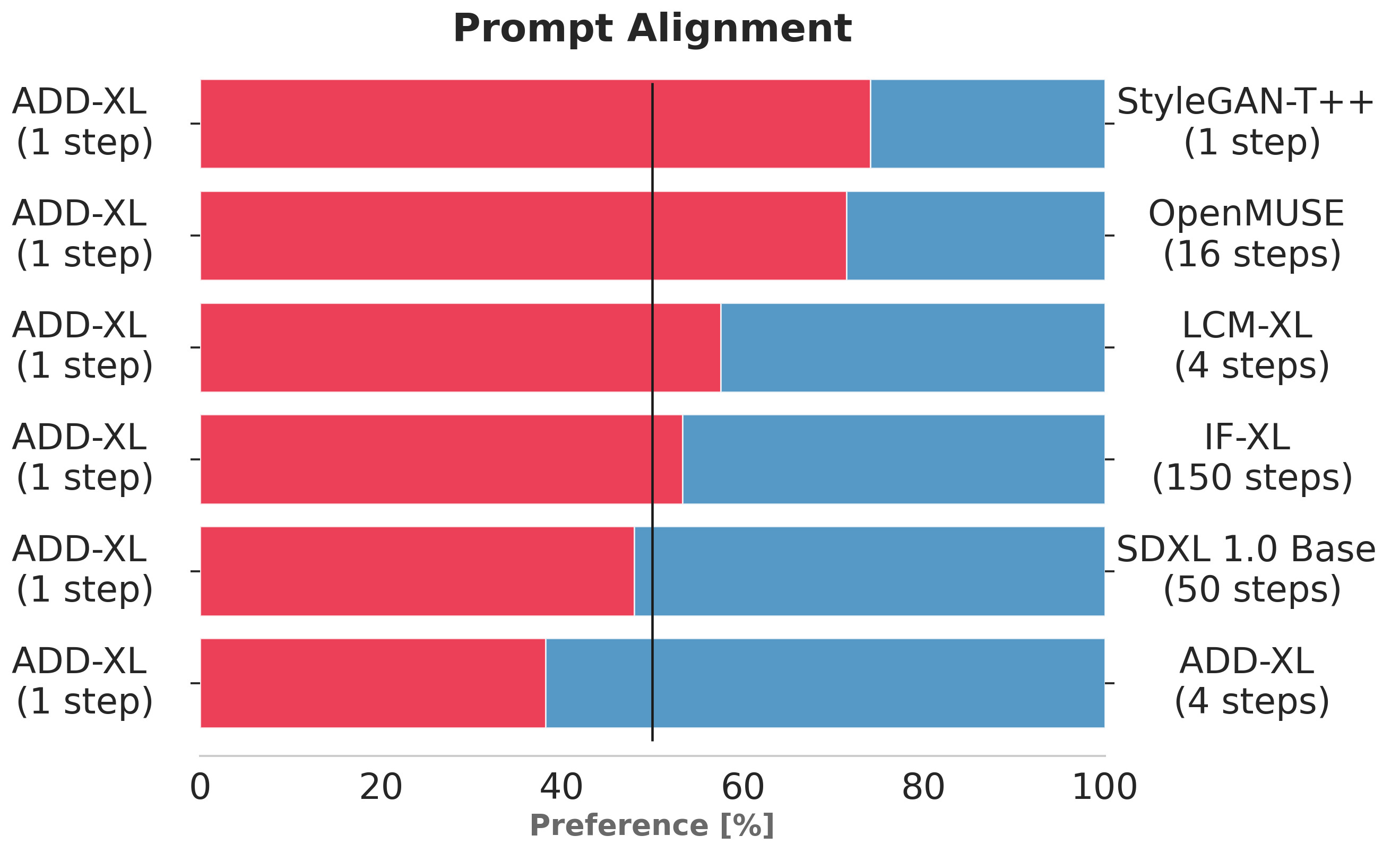}
\end{center}}\end{minipage}
}
\caption{
\textbf{User preference study (\textit{single step}).}
We compare the performance of \modelshort-XL (1-step) against established baselines.
\modelshort-XL model outperforms all models, except SDXL in human preference for both image quality and prompt alignment.
Using more sampling steps further improves our model (bottom row).
}
\label{fig:humanevalsingle}
\end{figure*}
}

\newcommand{\humanevalallmultiple}{
\begin{figure*}[t]
{
\centering
\begin{minipage}{0.49\linewidth}{
\begin{center}
\includegraphics[width=\linewidth]{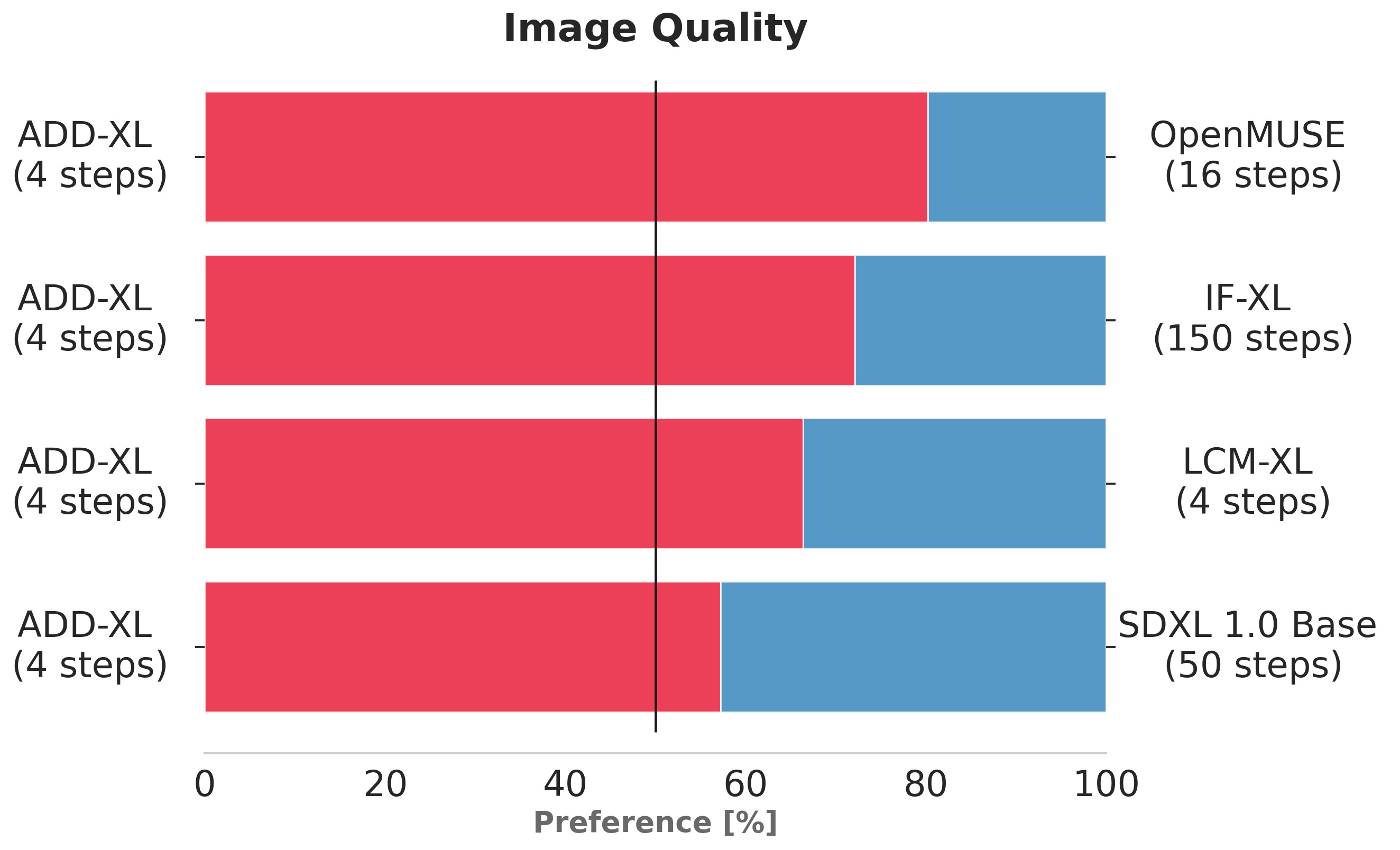}
\end{center}}\end{minipage}
}
\hfill
{
\centering
\begin{minipage}{0.49\linewidth}{
\begin{center}
\includegraphics[width=\linewidth]{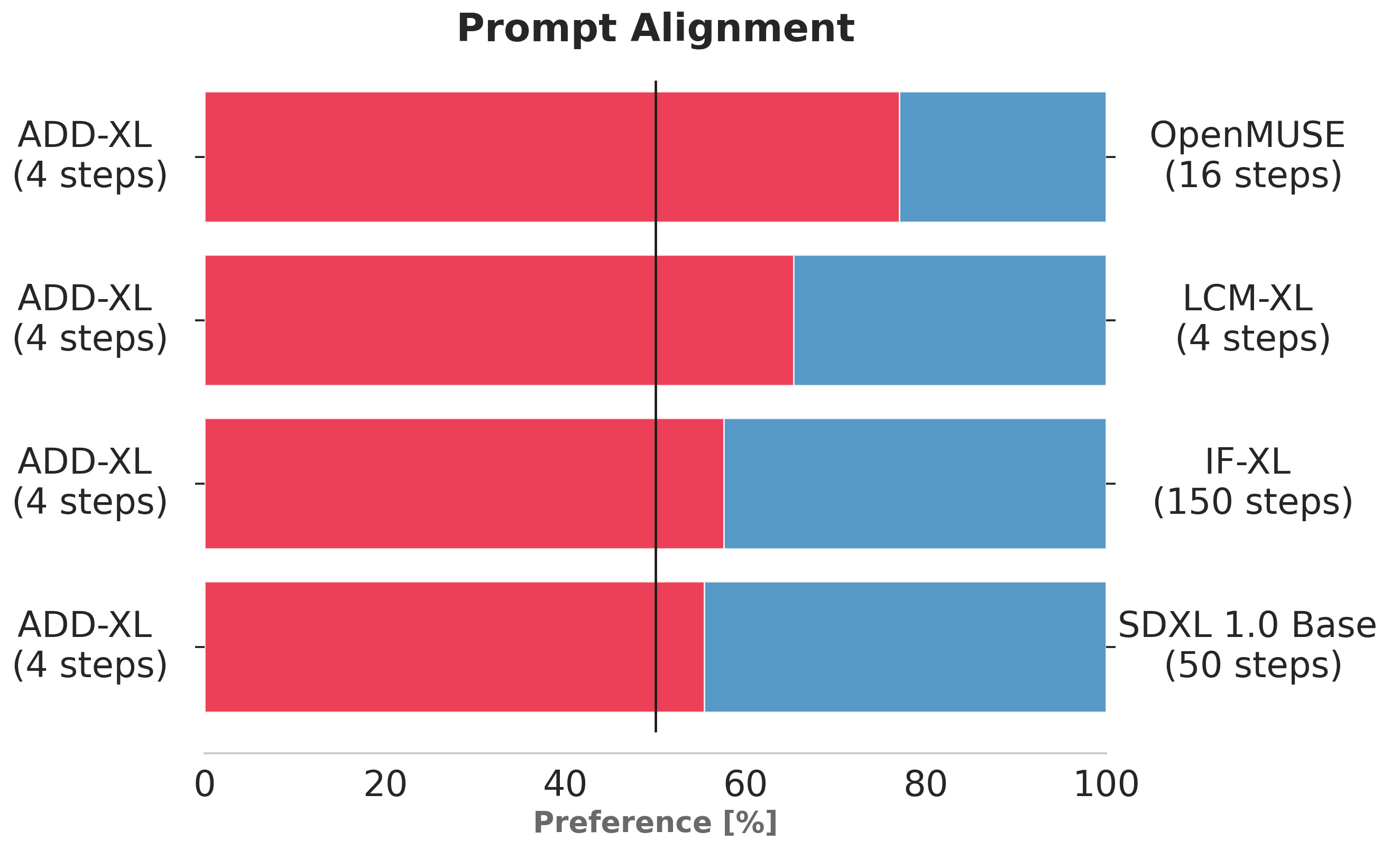}
\end{center}}\end{minipage}
}
\caption{
\textbf{User preference study (\textit{multiple steps}).}
We compare the performance of \modelshort-XL (4-step) against established baselines.
Our \modelshort-XL model outperforms all models, including its teacher SDXL 1.0 (base, no refiner)~\citep{podell2023sdxl}, in human preference for both image quality and prompt alignment.
}
\label{fig:humanevalmultiple}
\end{figure*}
}

\newcommand{\humanevalallsingleADDM}{
\begin{figure*}[h]
{
\centering
\begin{minipage}{0.49\linewidth}{
\begin{center}
\includegraphics[width=\linewidth]{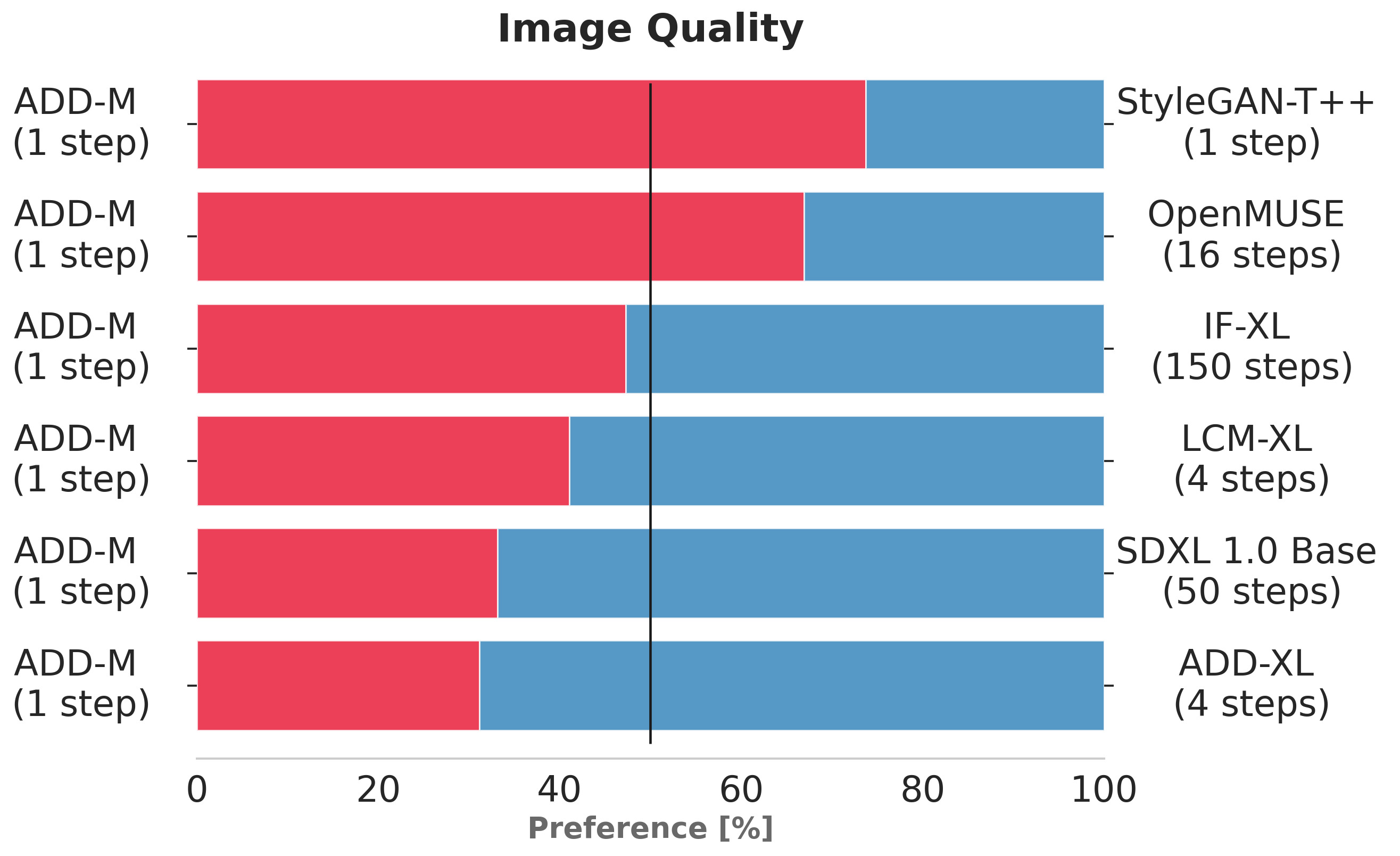}
\end{center}}\end{minipage}
}
\hfill
{
\begin{minipage}{0.49\linewidth}{
\begin{center}
\includegraphics[width=\linewidth]{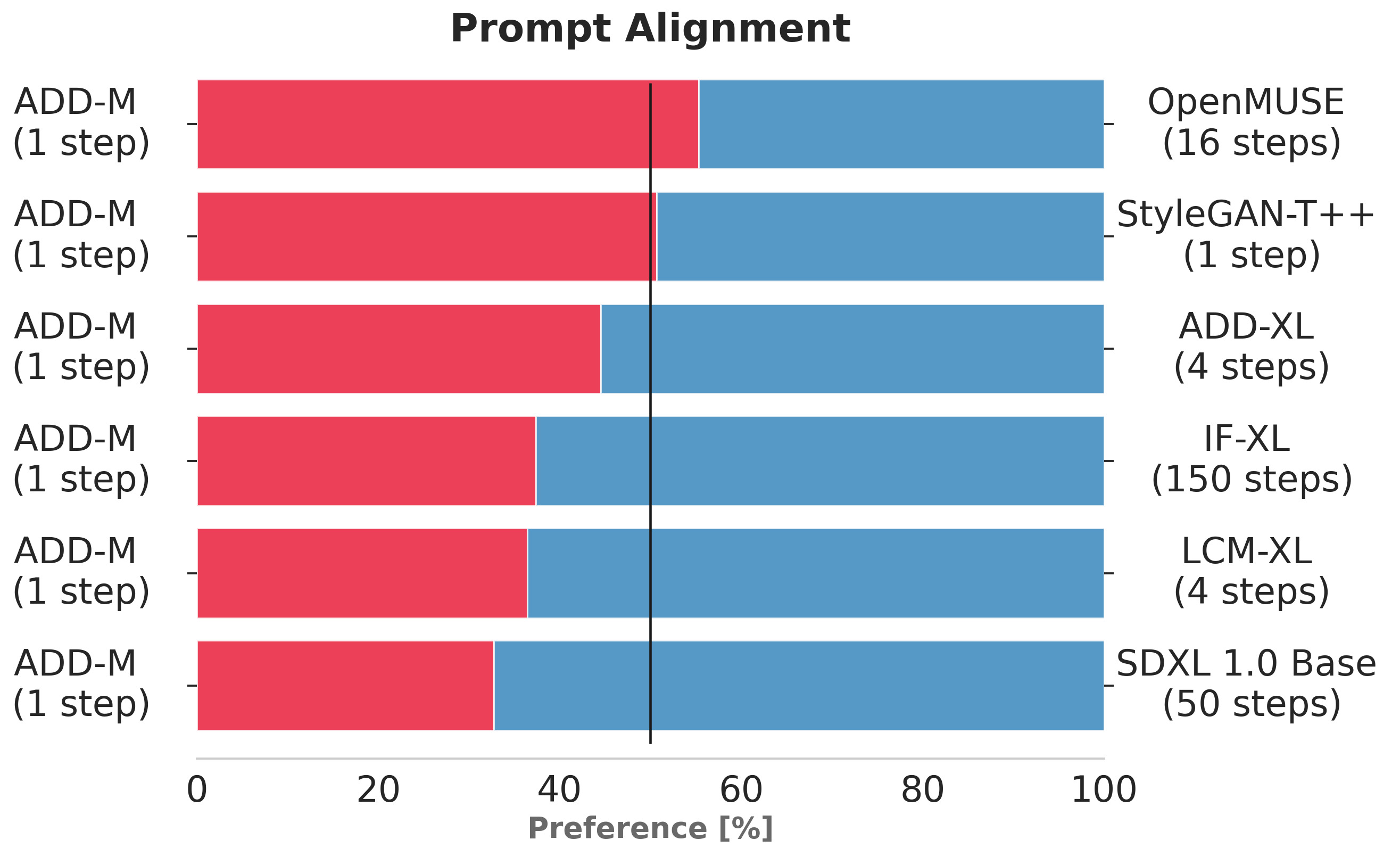}
\end{center}}\end{minipage}
}
\caption{
\textbf{User preference study (\textit{single step}).}
We compare the performance of \modelshort-M (1-step) against established baselines.
}
\label{fig:humanevalallsingleADDM}
\end{figure*}
}

\newcommand{\humanevalallmultipleADDM}{
\begin{figure*}[h]
{
\centering
\begin{minipage}{0.49\linewidth}{
\begin{center}
\includegraphics[width=\linewidth]{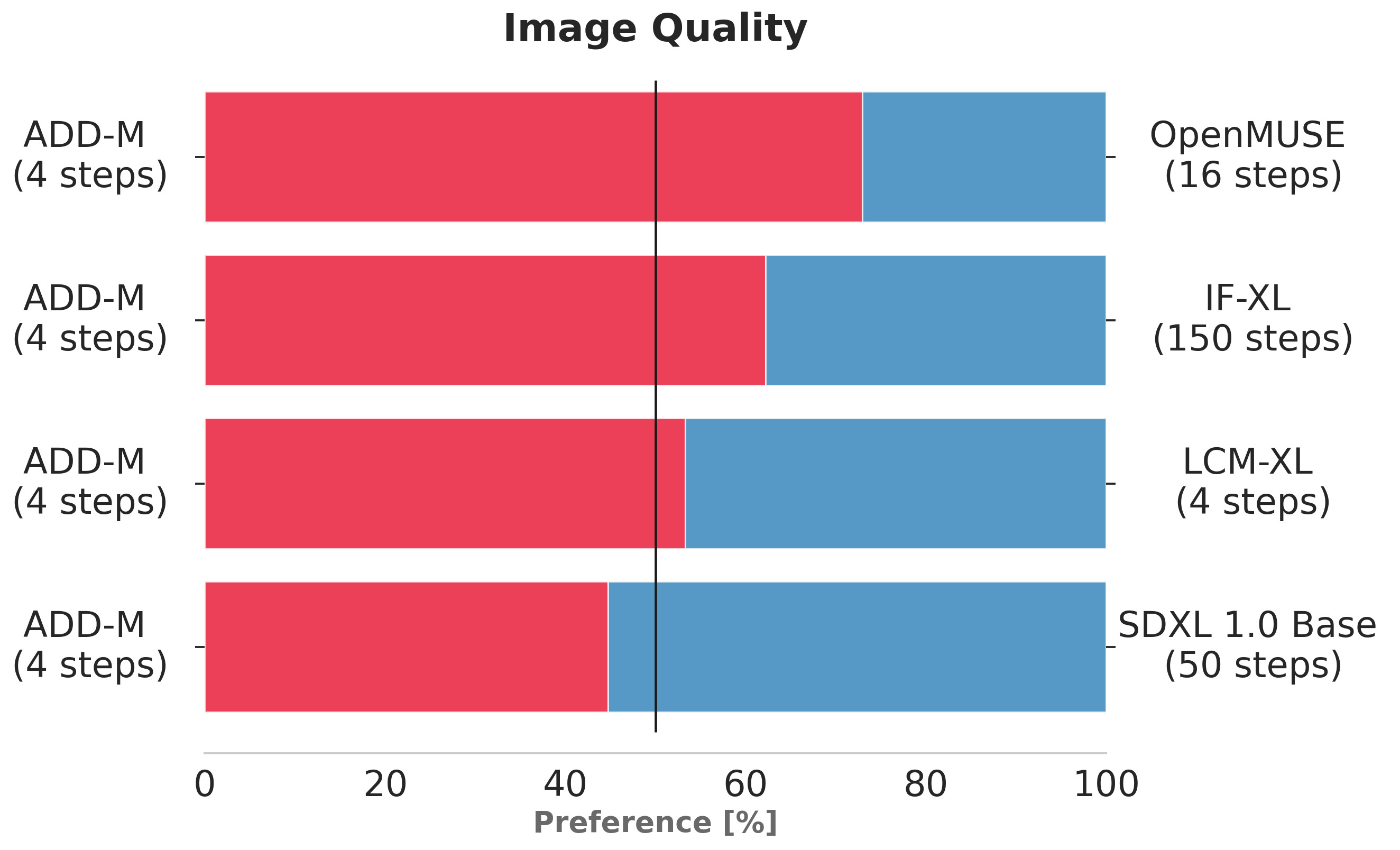}
\end{center}}\end{minipage}
}
\hfill
{
\centering
\begin{minipage}{0.49\linewidth}{
\begin{center}
\includegraphics[width=\linewidth]{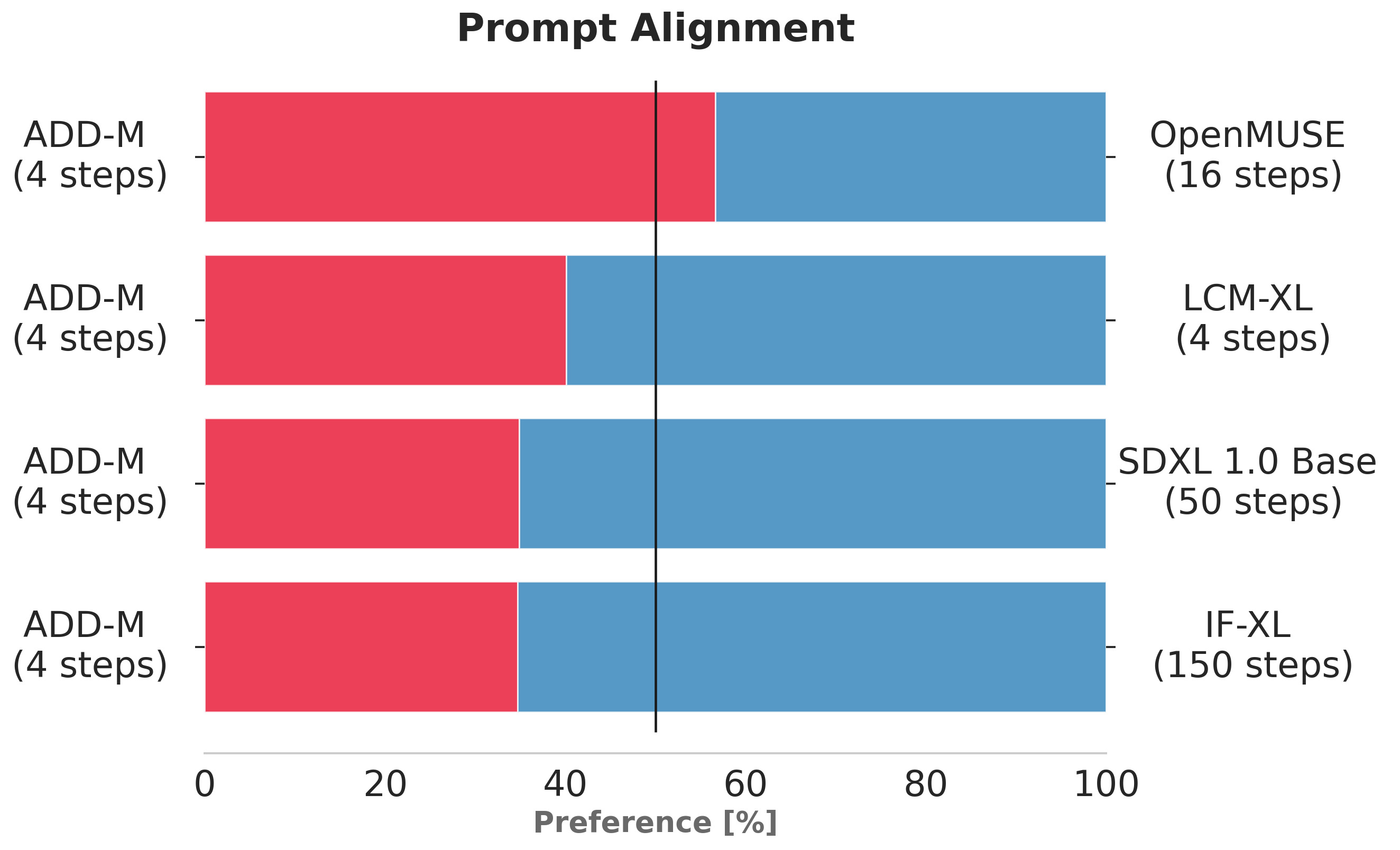}
\end{center}}\end{minipage}
}
\caption{
\textbf{User preference study (\textit{multiple steps}).}
We compare the performance of \modelshort-XL (4-step) against established baselines.
}
\label{fig:humanevalallmultipleADDM}
\end{figure*}
}

\newcommand{\figqualitativesteps}{
  \begin{figure*}[htbp]
    \centering
    \small
\begin{tabular}{llll}
    &
    \parbox[b]{.3\linewidth}{\centering ``A brain riding a rocketship heading towards the moon.''}
    \vspace{0.5em}
    &
    \parbox[b]{.3\linewidth}{\centering ``A bald eagle made of chocolate powder, mango, and whipped cream''}
    &
    \parbox[b]{.3\linewidth}{\centering ``A blue colored dog.''}
    \\
  \rotatebox[origin=c]{90}{1 step} &
  \includegraphics[width=.3\linewidth,valign=m]{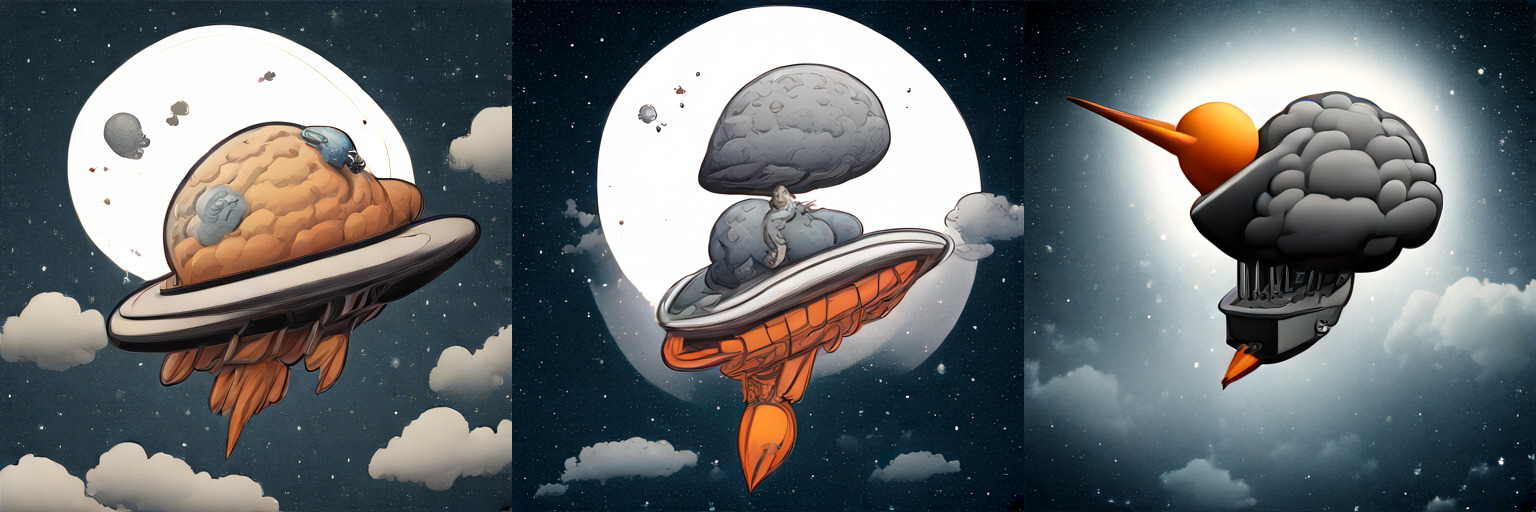}
  &
  \includegraphics[width=.3\linewidth,valign=m]{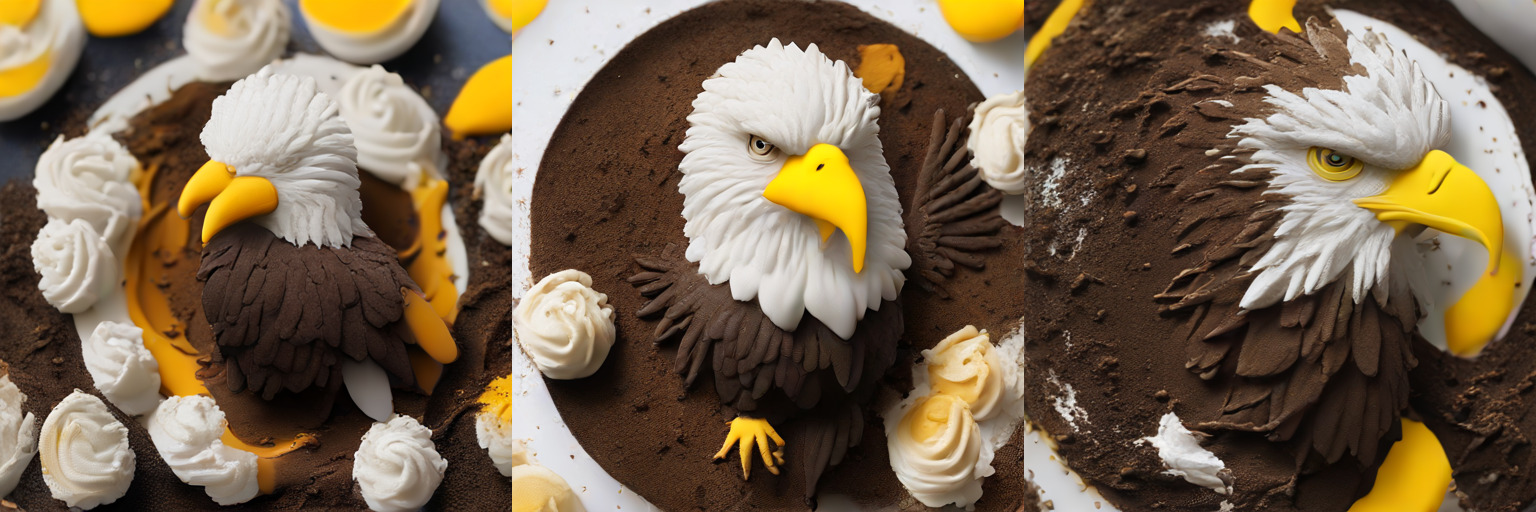}
  &
  \includegraphics[width=.3\linewidth,valign=m]{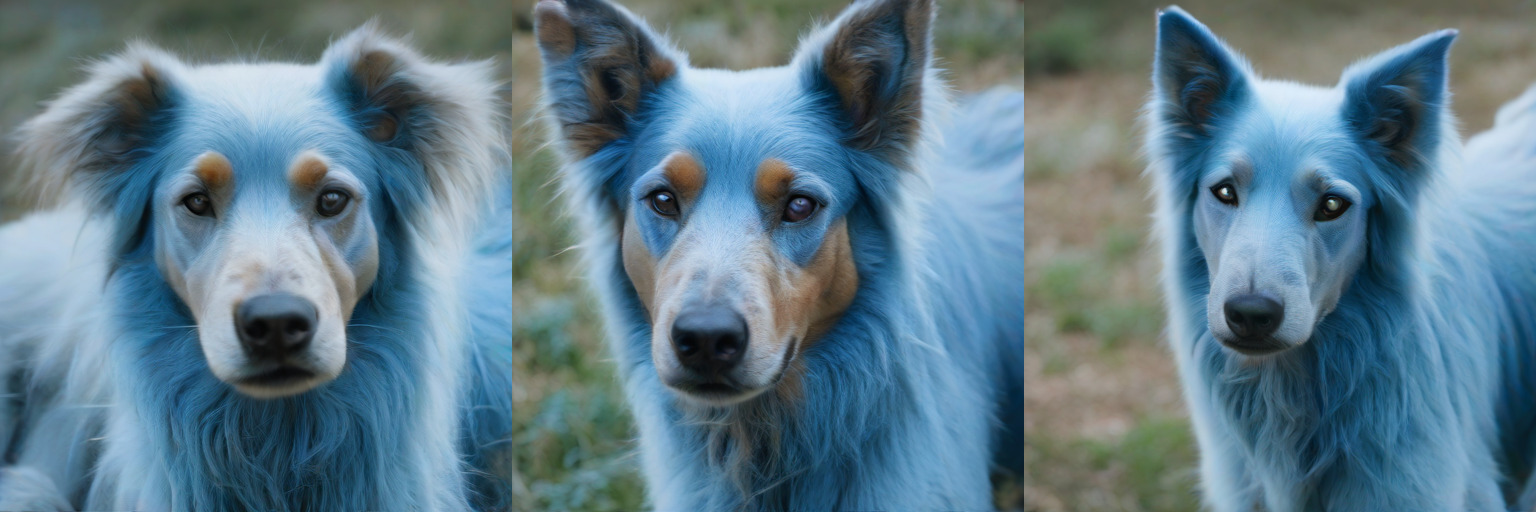}\\
  \rotatebox[origin=c]{90}{2 steps} &
  \includegraphics[width=.3\linewidth,valign=m]{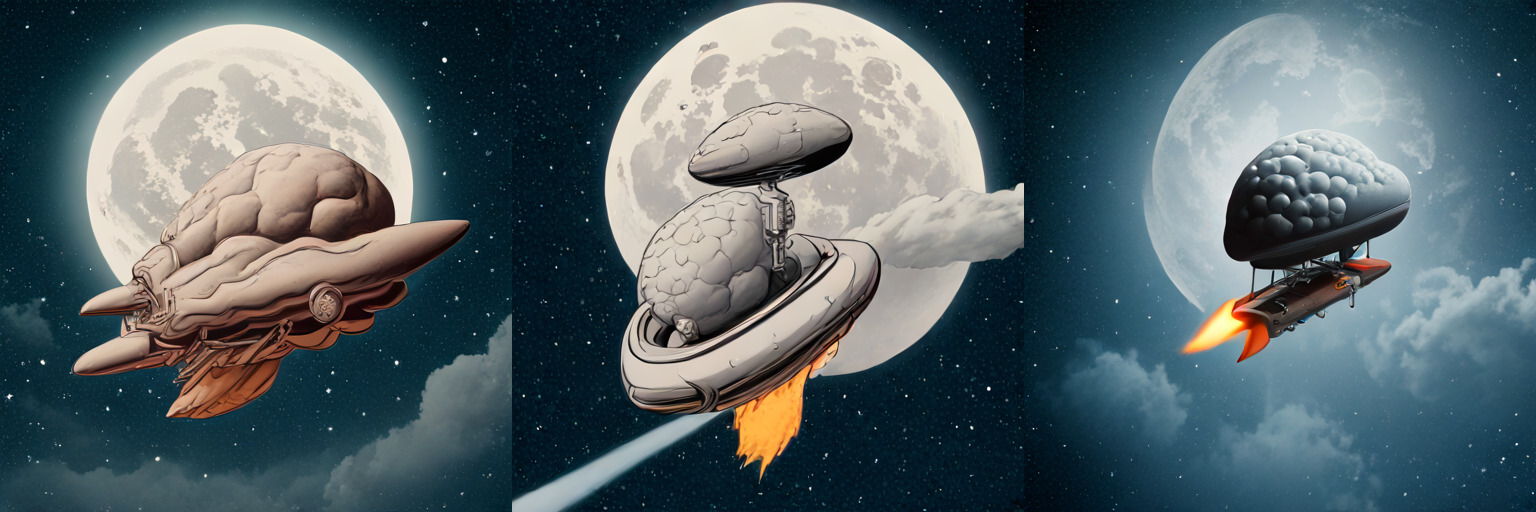}
  &
  \includegraphics[width=.3\linewidth,valign=m]{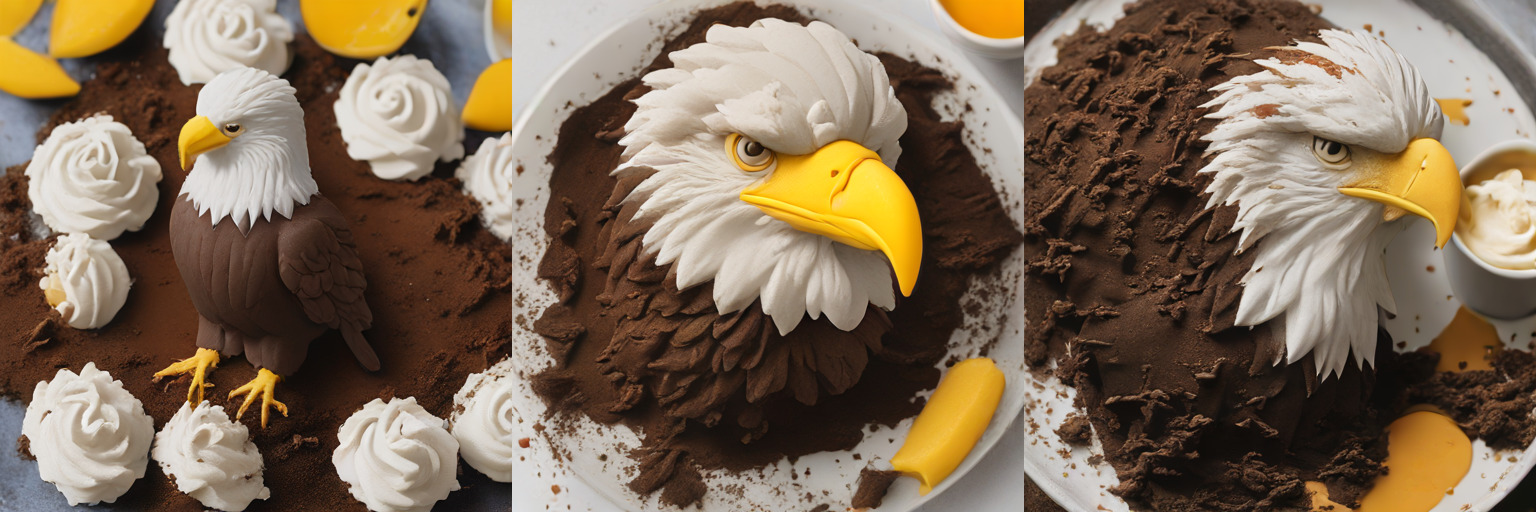}
  &
  \includegraphics[width=.3\linewidth,valign=m]{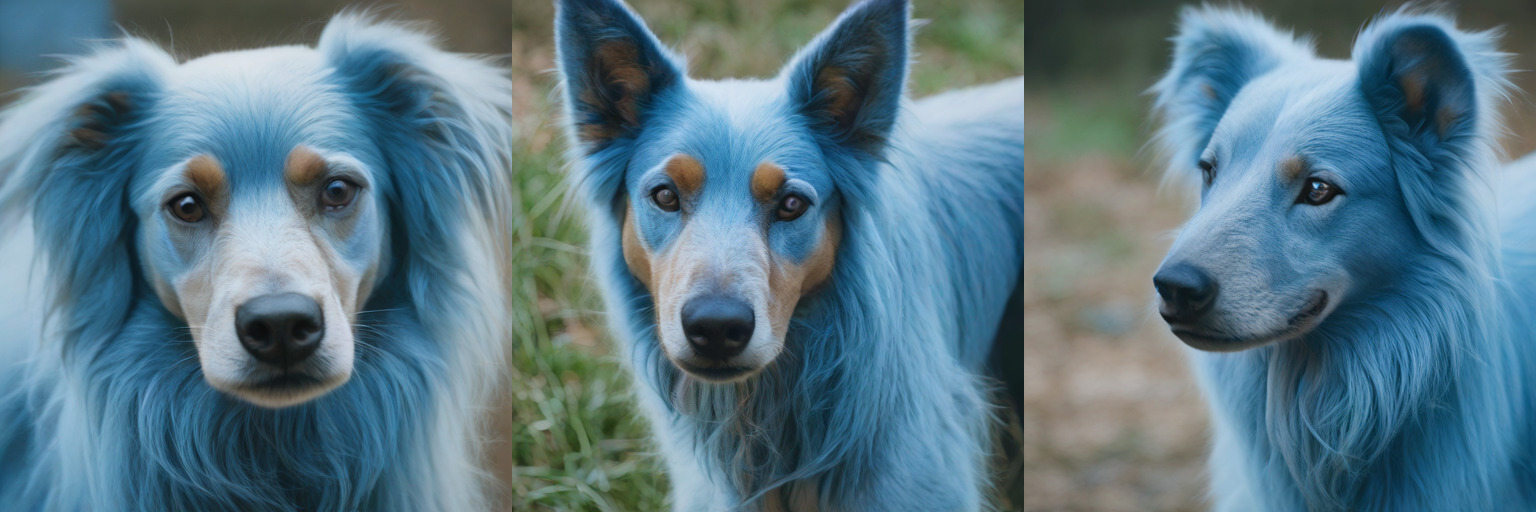}\\
  \rotatebox[origin=c]{90}{4 steps} &
  \includegraphics[width=.3\linewidth,valign=m]{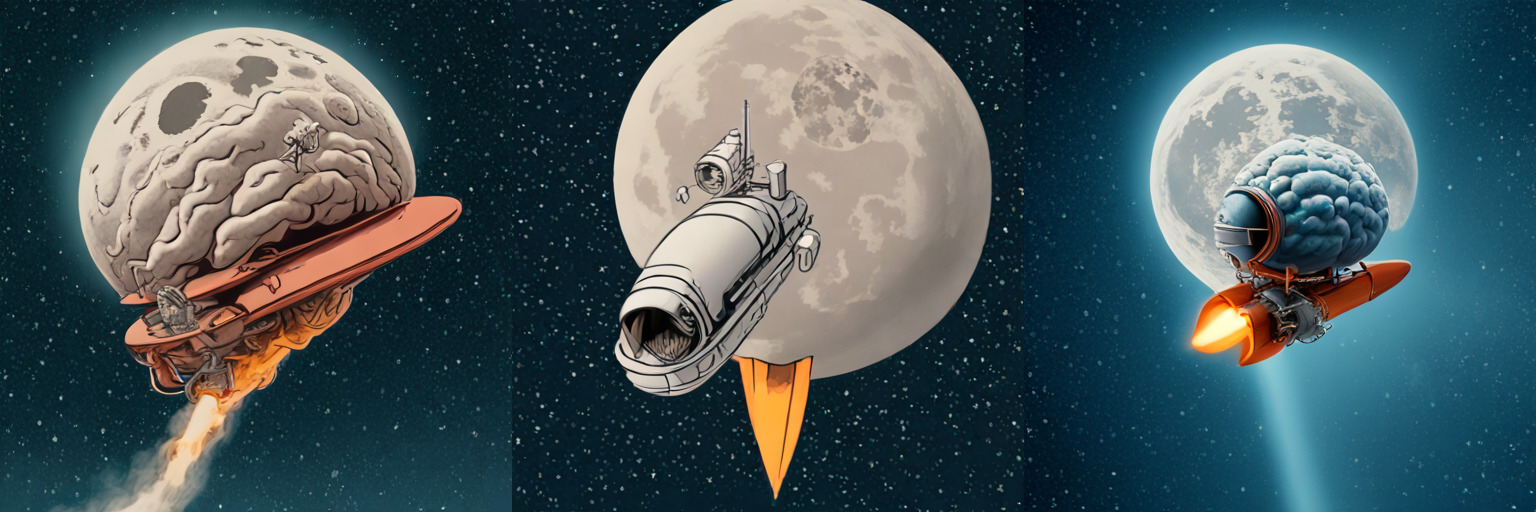}
  &
  \includegraphics[width=.3\linewidth,valign=m]{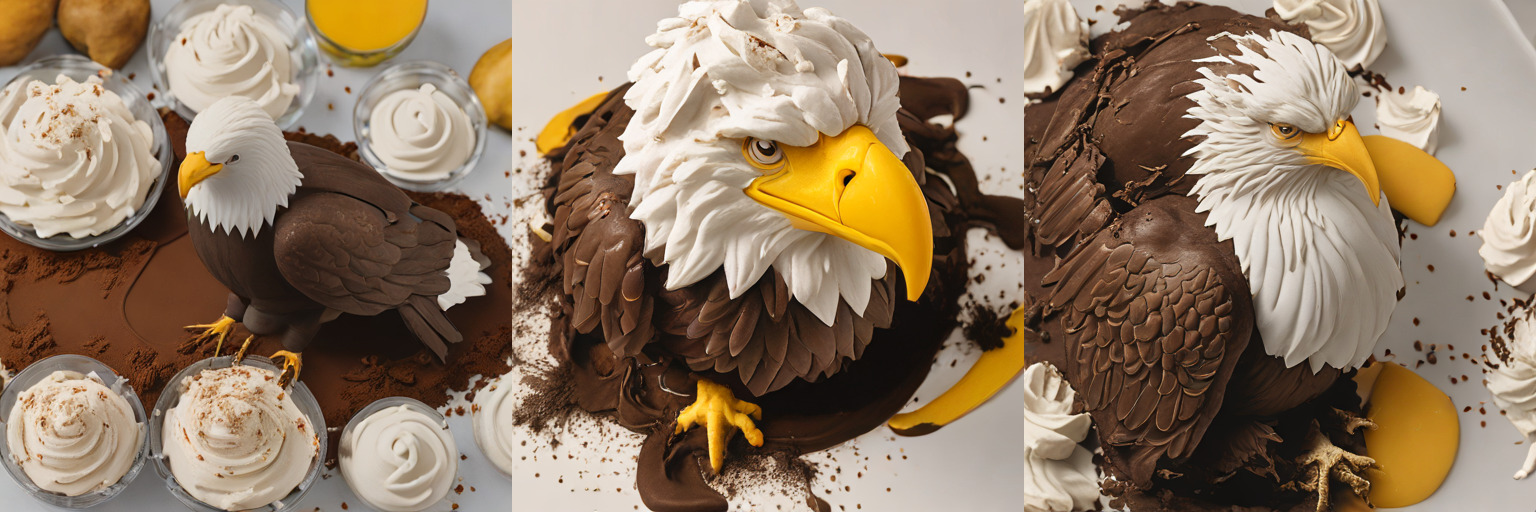}
  &
  \includegraphics[width=.3\linewidth,valign=m]{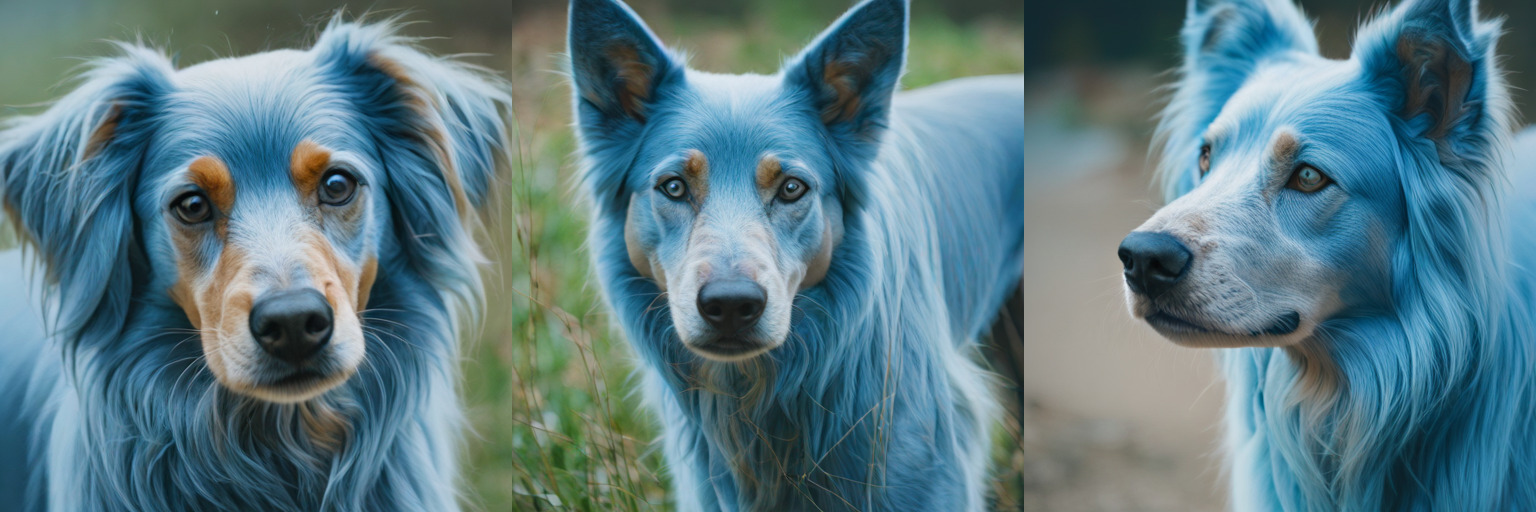}\\
\end{tabular}
    \caption{\textbf{Qualitative effect of sampling steps.} We show qualitative examples
    when sampling \modelshort-XL with 1, 2, and 4 steps. Single-step samples are often already of high quality, but increasing the number of steps can
    further improve the consistency
    (\eg second prompt, first column) and attention to detail (\eg second prompt, second column). 
    The seeds are constant within columns and we see that the general layout is preserved across sampling steps, allowing for fast exploration of
    outputs while retaining the possibility to refine.
    \vspace{-1em}
    \label{fig:qualitativesteps}
    }
  \end{figure*}
}

\newcommand{\truncationstylegant}{
(0.304709415435791, 24.232327533869036)
(0.30354503631591795, 23.214245802031257)
(0.3020538139343262, 22.26808578751586)
(0.30178037643432615, 21.964200338343055)
(0.2987831497192383, 20.980434378225244)
(0.2983003044128418, 20.573519219463595)
(0.2971324348449707, 20.13091776414437)
(0.2945480537414551, 19.795256875788045)
(0.2928913116455078, 19.351840066034253)
}

\newcommand{\truncationgigagan}{
(0.321, 15.4)
(0.325, 17.4)
(0.327, 19.8)
(0.329, 21.9)
}

\newcommand{\truncationstylegantplusplus}{
(0.3441916406154632, 16.32707192884507)
(0.3448771834373474, 16.46717814180094)
(0.3461066782474518, 16.885803464262956)
(0.3475224971771240, 17.18813963803478)
}

\newcommand{\truncationCurve}[3]{ %
  \addplot[#1, very thick, mark=*, mark size=1.2pt] coordinates {#2};
  \addlegendentry{\hspace{0.5mm}#3};
}

\newcommand{\truncation}{
\begin{figure}[htbp]
\centering%
\resizebox{\linewidth}{!}{%
\begin{tikzpicture}%
\begin{axis}[
  width=100mm, height=45mm,
  xlabel={CLIP score (ViT-g-14)}, xmin={0.28}, xmax={0.36}, xmode={linear}, xtick={0.29,0.30,0.31,0.32,0.33,0.34,0.35,0.36},
  ylabel={Zero-shot FID\textsubscript{5k}}, ymin={14}, ymax={26}, ymode={linear}, ytick={14, 16, 18, 20, 22, 24, 26},
  grid={major}, 
  legend pos={north east}, legend cell align={left},
  legend style={nodes={scale=0.5, transform shape}}, legend image post style={mark=*}]
]
\truncationCurve{C0}{\truncationgigagan}{GigaGAN}
\truncationCurve{C1}{\truncationstylegant}{StyleGAN-T}
\truncationCurve{C2}{\truncationstylegantplusplus}{StyleGAN-T++}
\end{axis}
\end{tikzpicture}%
}%
\vspace*{-2mm}%
\caption{ 
\textbf{Comparing text alignment tradeoffs at 256 $\times$ 256 pixels.} 
We compare FID--CLIP score curves of StyleGAN-T, StyleGAN-T++, and GigaGAN.
For increasing CLIP score, all methods use via decreasing truncation~\cite{Karras2018ASG} for values $\psi=\{1.0, 0.9, \dots, 0.3\}$.
}
\label{fig:truncation}
\end{figure}
}

\newcommand{\suppteaser}{
\vspace{3em}
\begin{figure}[htbp]
\includegraphics[width=\textwidth]{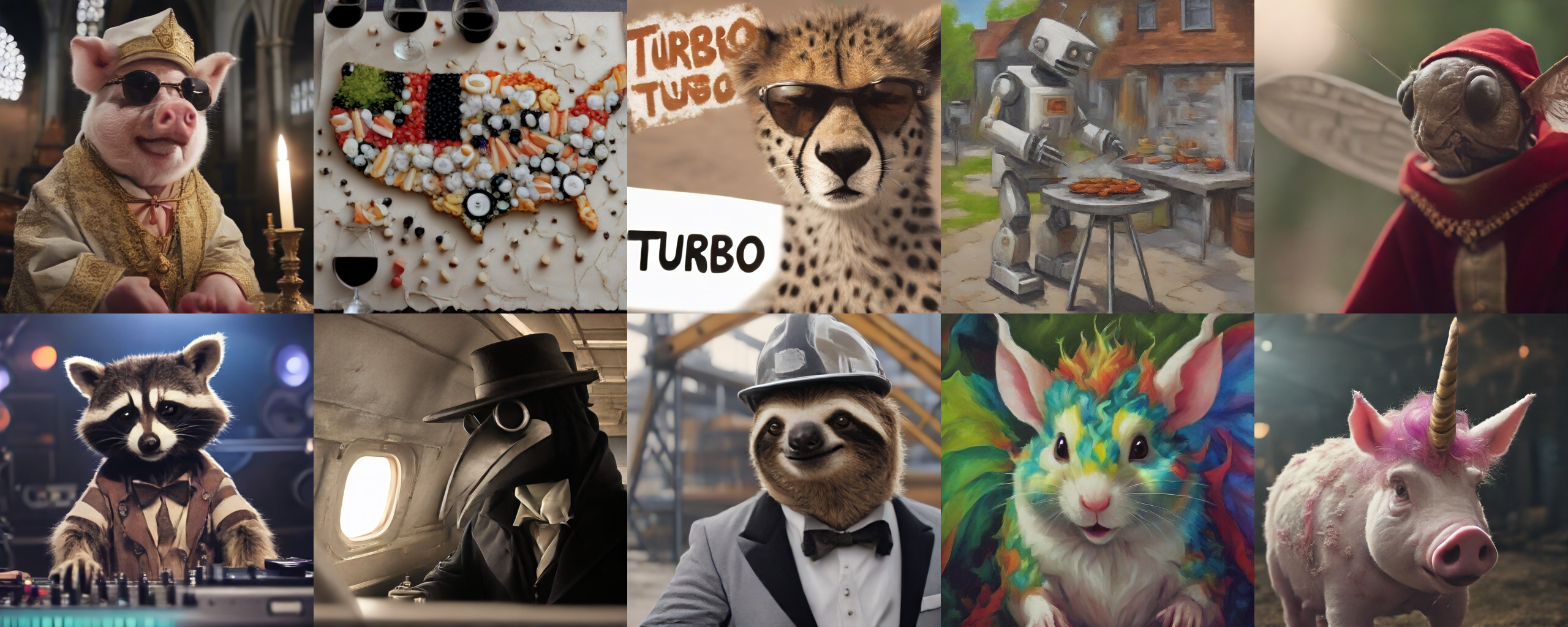}
\caption{\label{fig:suppteaser}
\textbf{Additional single step $512^2$ images generated with \modelshort-XL.}
All samples are generated with a single U-Net evaluation trained with adversarial diffusion distillation (ADD).
}
\end{figure}
}

\newcommand{\suppsteps}{
  \begin{figure*}[htbp]
    \centering
    \small
\begin{tabular}{lll}
    &
    \parbox[b]{.45\linewidth}{\centering ``a robot is playing the guitar at a rock concert in front of a large crowd.''}
    \vspace{0.5em}
    &
    \parbox[b]{.45\linewidth}{\centering ``A portrait photo of a kangaroo wearing an orange hoodie and blue sunglasses standing on the grass in front of the Sydney Opera House holding a sign on the chest that says Welcome Friends!''}
    \\
  \rotatebox[origin=c]{90}{1 step} &
  \includegraphics[width=.45\linewidth,valign=m]{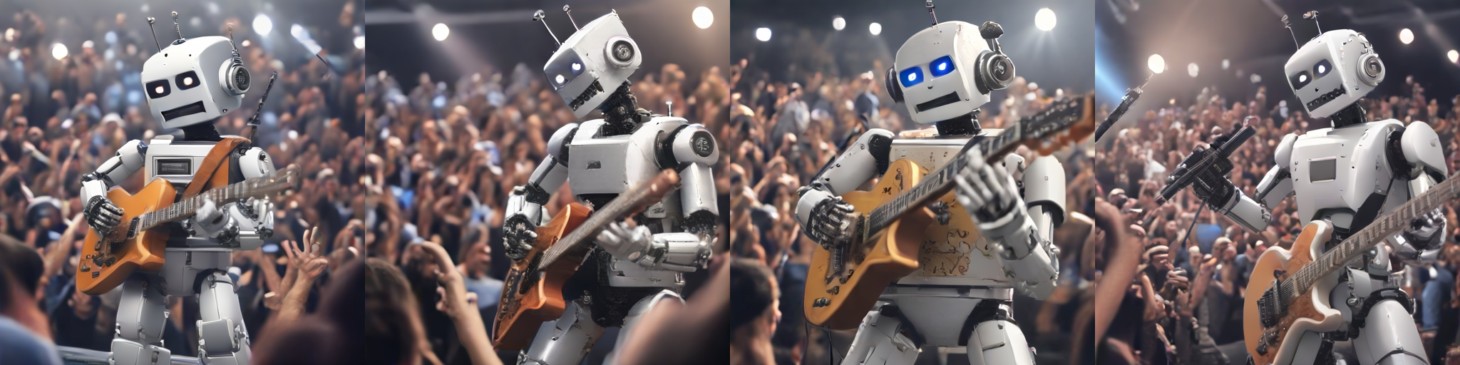}
  &
  \includegraphics[width=.45\linewidth,valign=m]{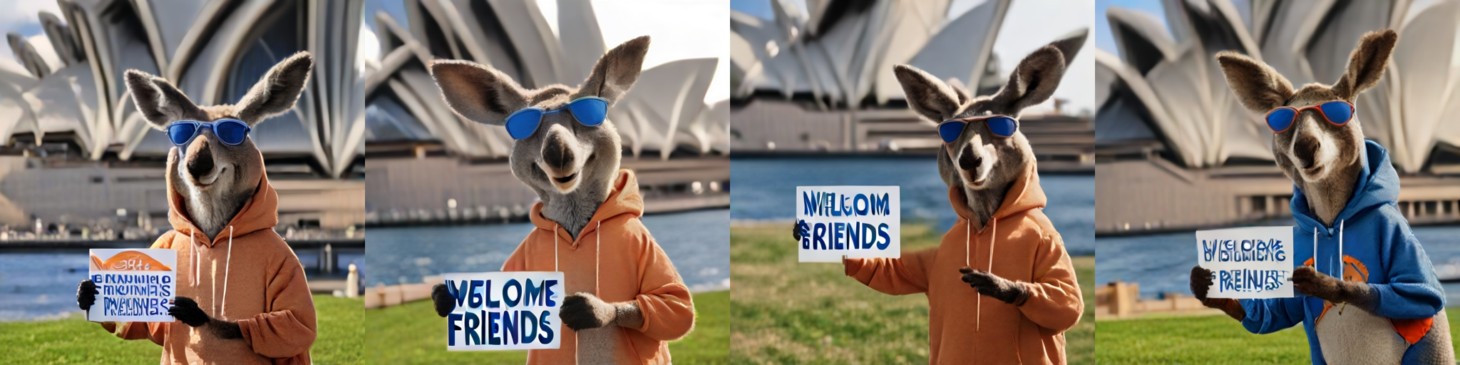}
  \\
  \rotatebox[origin=c]{90}{2 steps} &
  \includegraphics[width=.45\linewidth,valign=m]{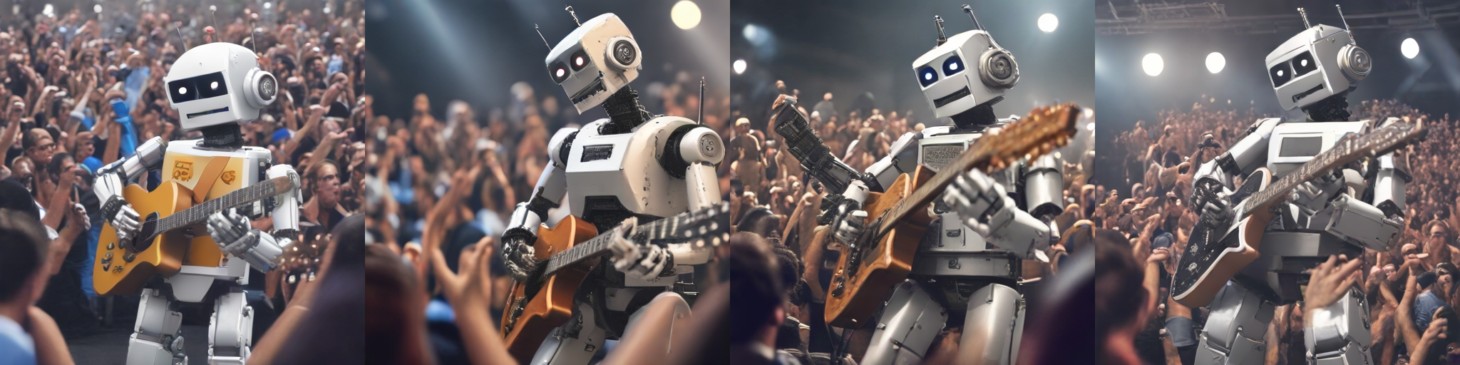}
  &
  \includegraphics[width=.45\linewidth,valign=m]{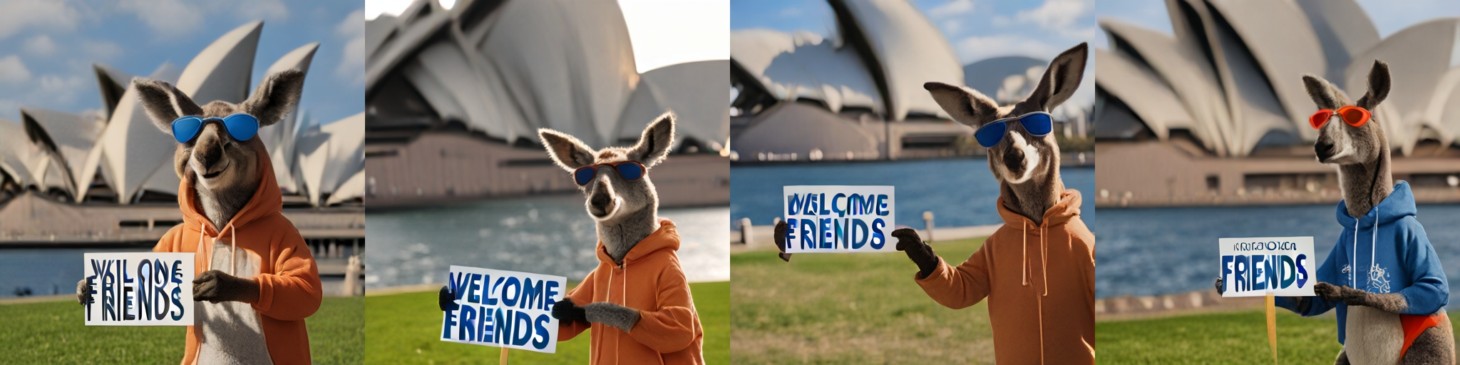}
  \\
  \rotatebox[origin=c]{90}{4 steps} &
  \includegraphics[width=.45\linewidth,valign=m]{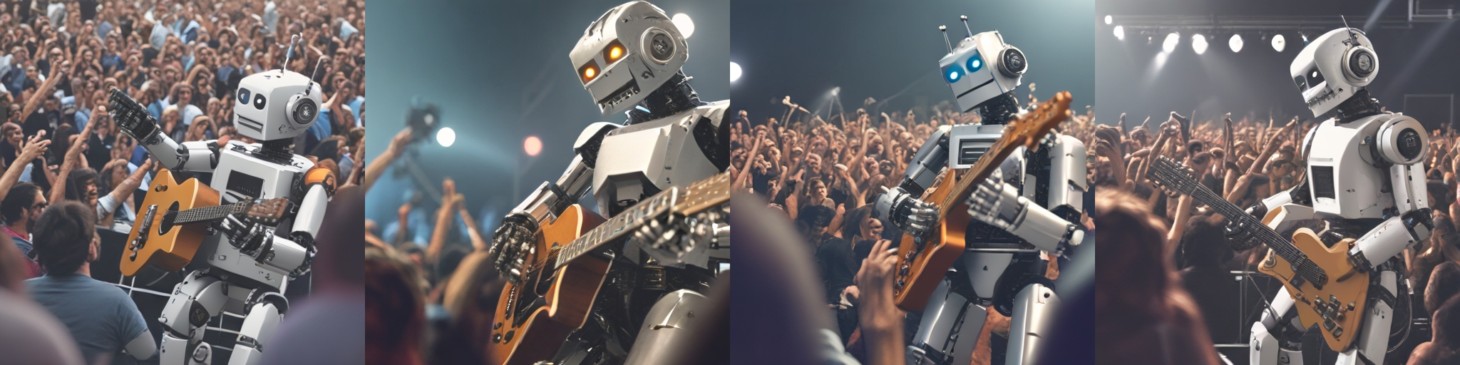}
  &
  \includegraphics[width=.45\linewidth,valign=m]{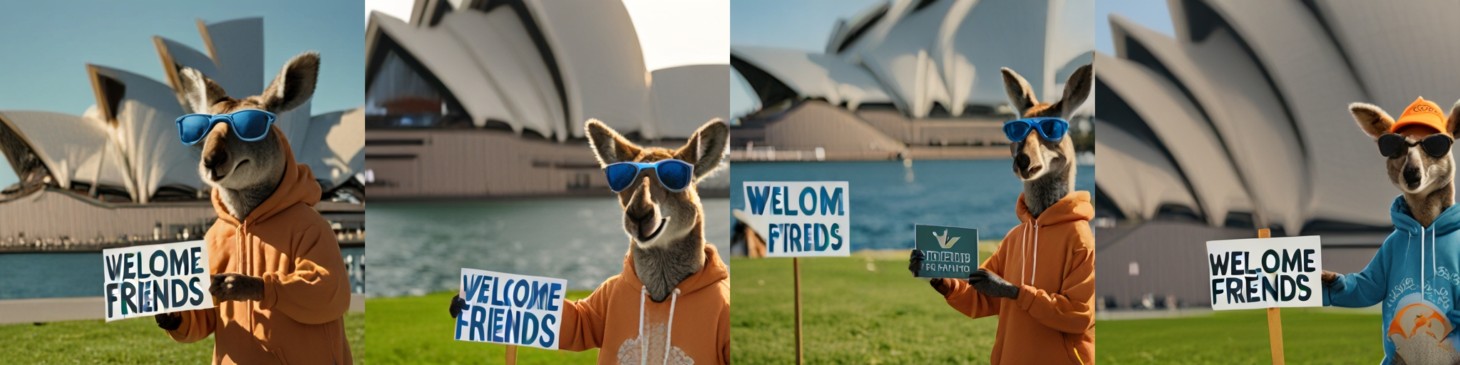}
  \\
\end{tabular}
    \caption{\textbf{Additional results on the qualitative effect of sampling steps.} Similar to \Cref{fig:qualitativesteps}, we show qualitative examples
    when sampling \modelshort-XL with 1, 2, and 4 steps. 
    Single-step samples are often already of high quality, but increasing the number of steps can further improve the diversity
    (left) and spelling capabilities (right). 
    The seeds are constant within columns and we see that the general layout is preserved across sampling steps, allowing for fast exploration of
    outputs while retaining the possibility to refine.
    \label{fig:suppsteps}
    }
  \end{figure*}
}

\newcommand{\qualitativecompsupp}{
  \begin{figure*}[htbp]
    \centering
    \small
\begin{tabular}{lll}
    &
    \parbox[b]{.45\linewidth}{\centering \emph{A cinematic shot of robot with colorful feathers.}}
    \vspace{0.5em}
    &
    \parbox[b]{.45\linewidth}{\centering \emph{Teddy bears working on new AI research on the moon in the 1980s.}}
    \\
  \rotatebox[origin=c]{90}{\makecell{\modelshort-XL \\ (1 step)}} &
  \includegraphics[width=.45\linewidth,valign=m]{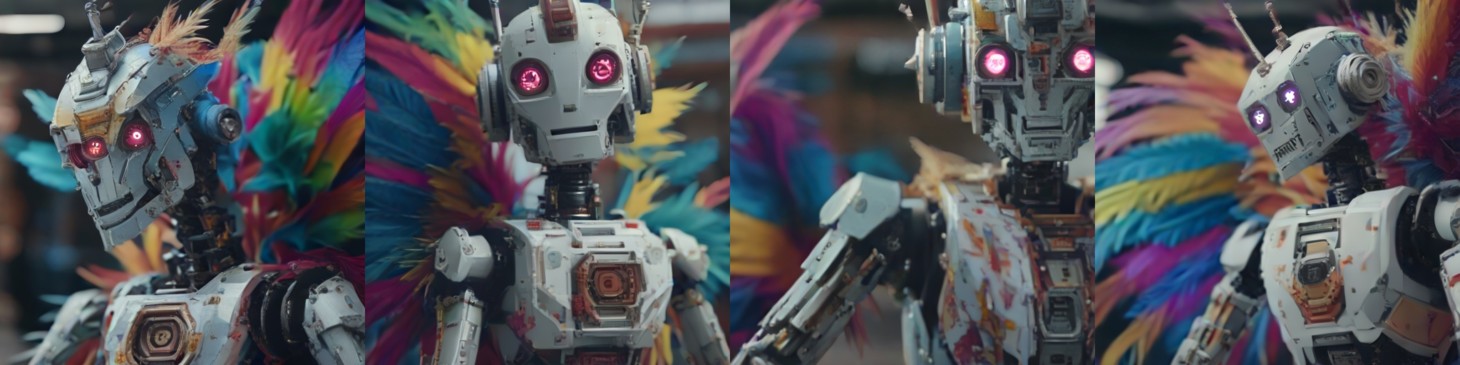}
  &
  \includegraphics[width=.45\linewidth,valign=m]{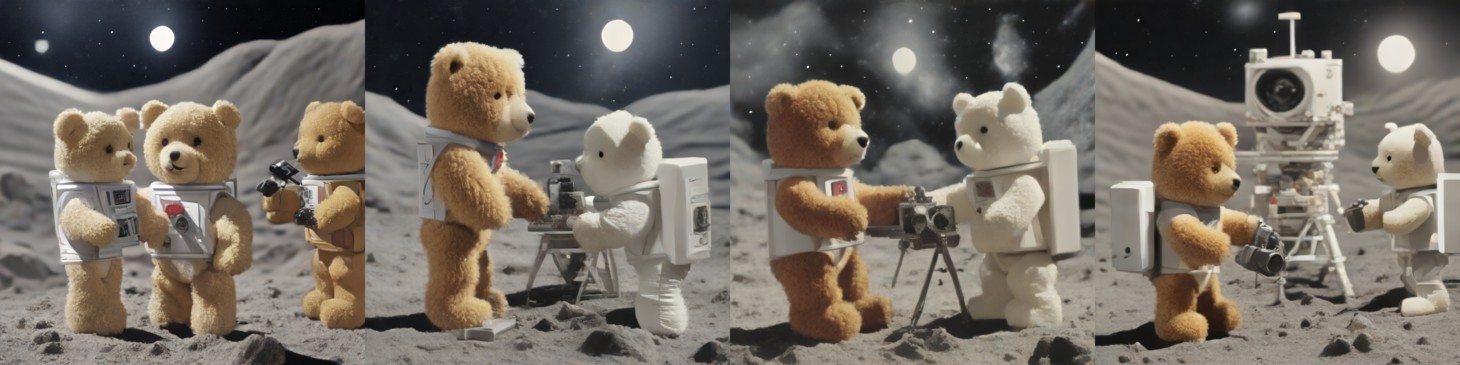}
  \\
  \rotatebox[origin=c]{90}{\makecell{LCM-XL \\ (1 step)}} &
  \includegraphics[width=.45\linewidth,valign=m]{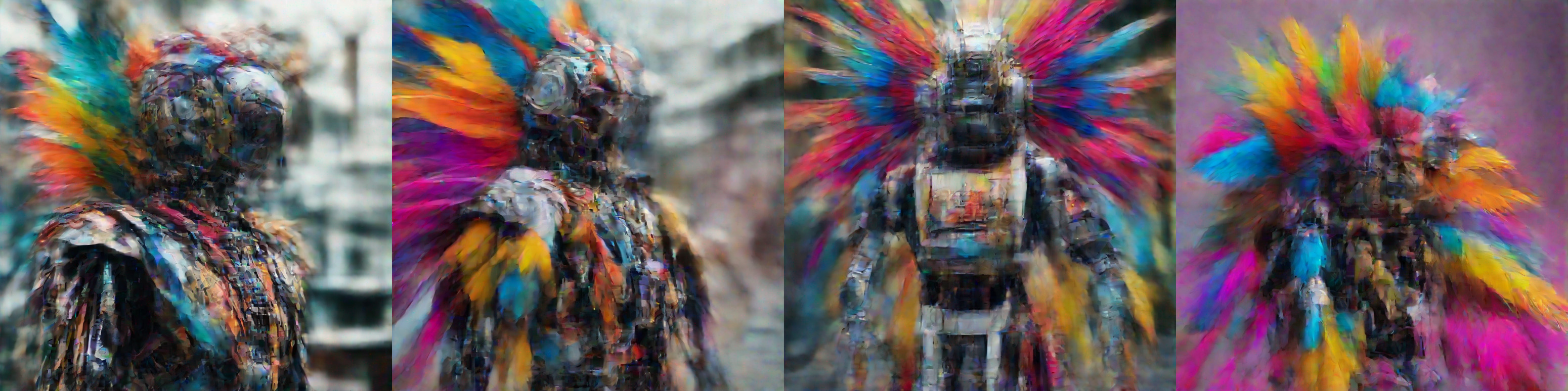}
  &
  \includegraphics[width=.45\linewidth,valign=m]{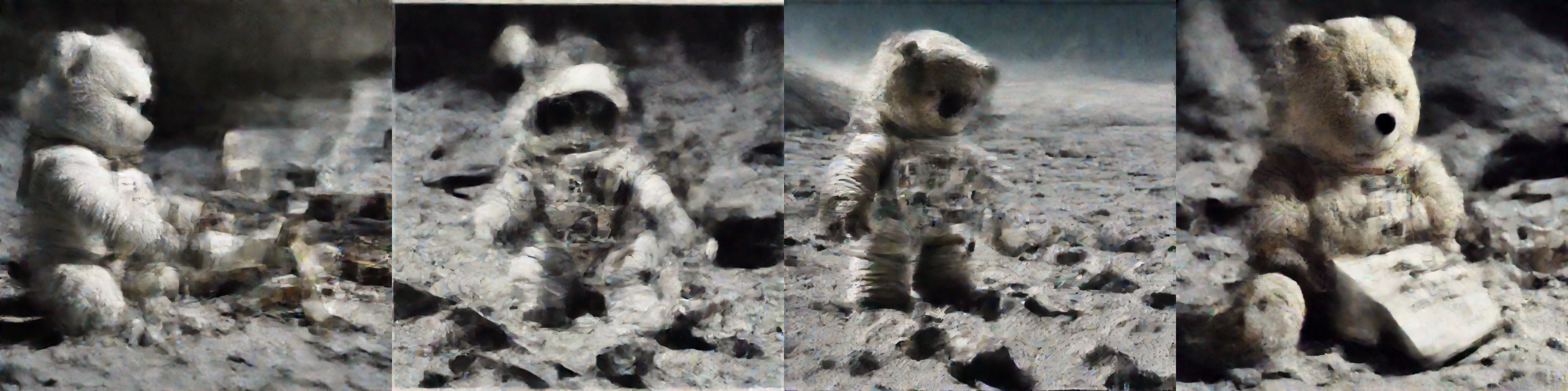}
  \\
  \rotatebox[origin=c]{90}{\makecell{\modelshort-XL \\ (2 steps)}} &
  \includegraphics[width=.45\linewidth,valign=m]{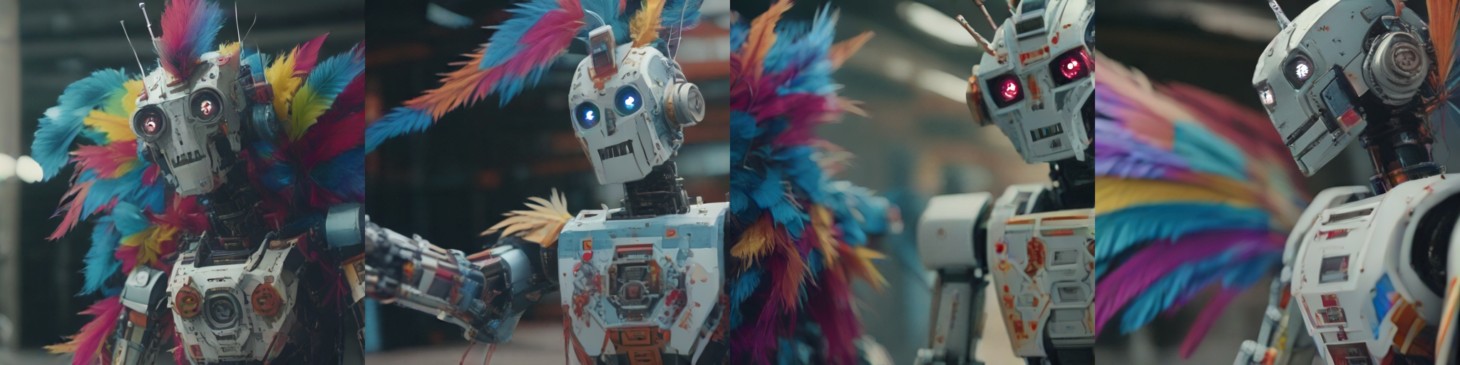}
  &
  \includegraphics[width=.45\linewidth,valign=m]{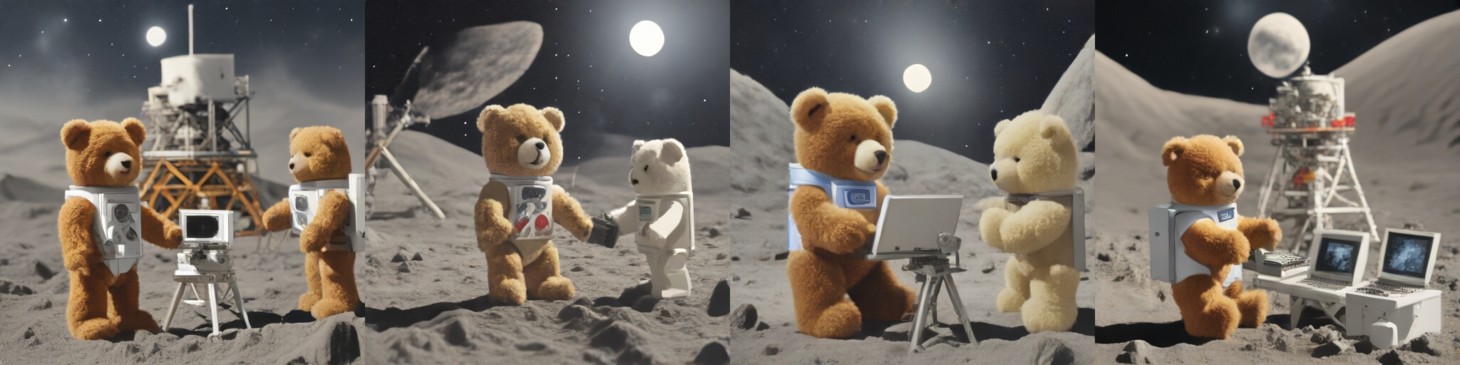}
  \\
  \rotatebox[origin=c]{90}{\makecell{LCM-XL \\ (2 steps)}} &
  \includegraphics[width=.45\linewidth,valign=m]{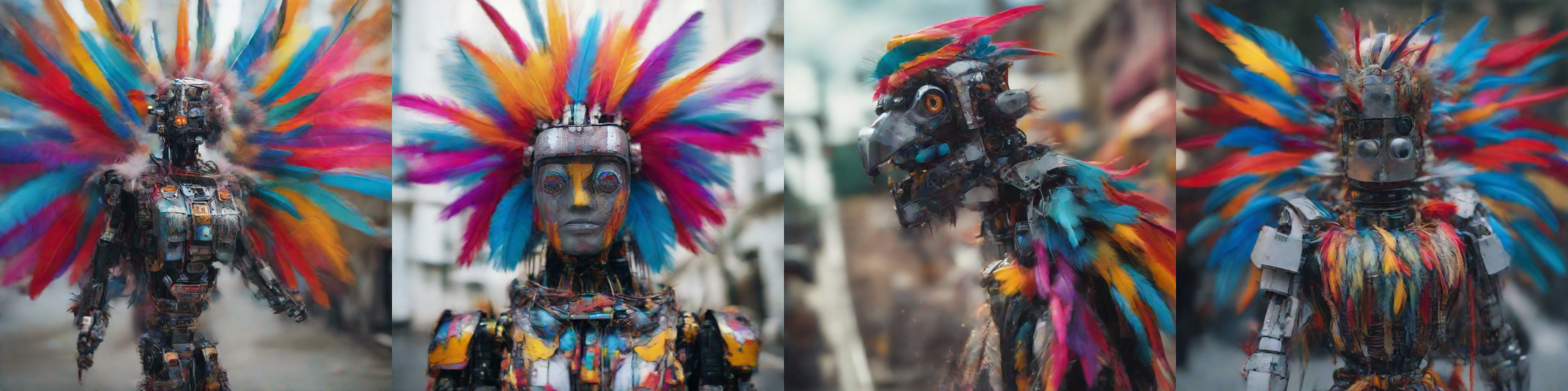}
  &
  \includegraphics[width=.45\linewidth,valign=m]{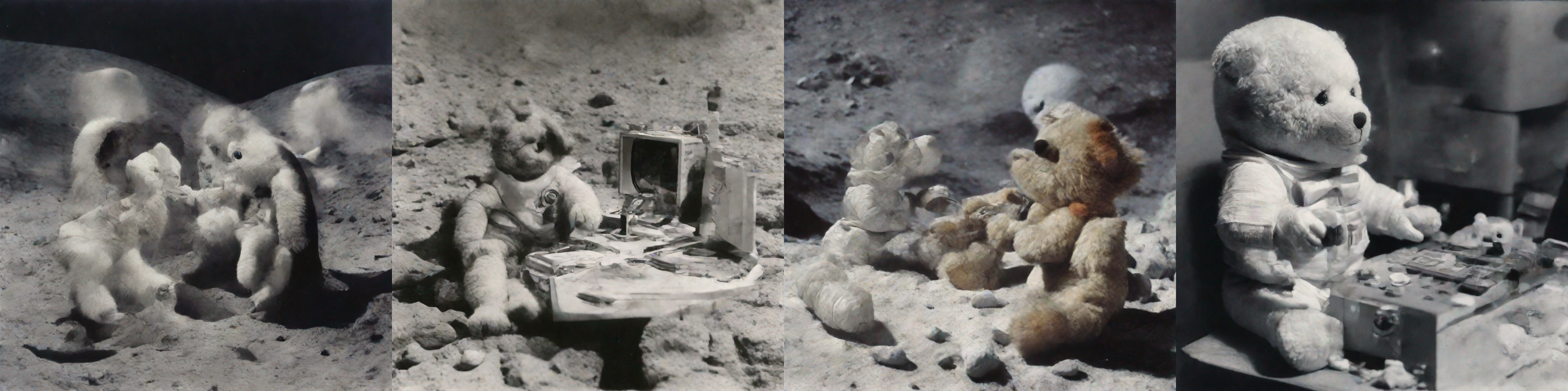}
  \\
  \rotatebox[origin=c]{90}{\makecell{\modelshort-XL \\ (4 steps)}} &
  \includegraphics[width=.45\linewidth,valign=m]{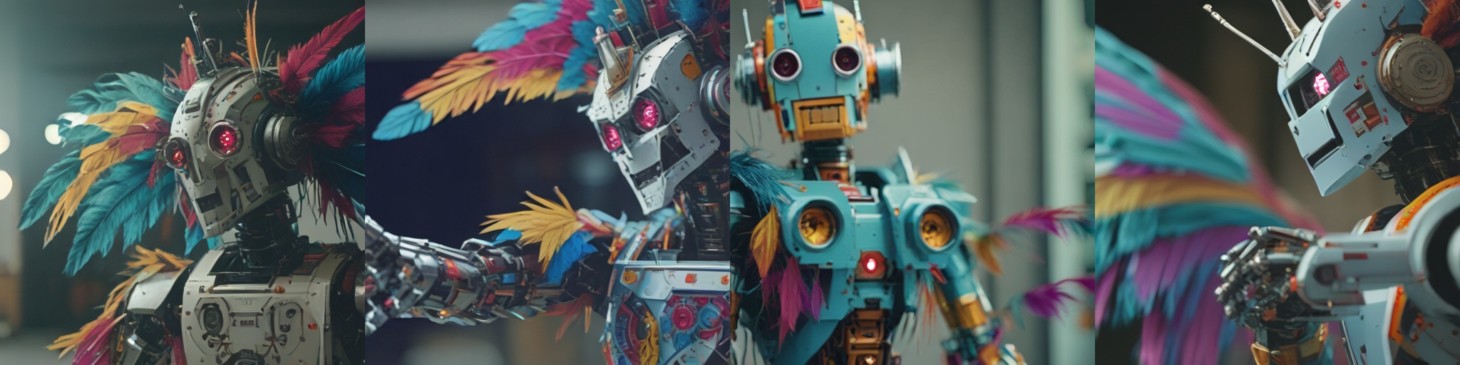}
  &
  \includegraphics[width=.45\linewidth,valign=m]{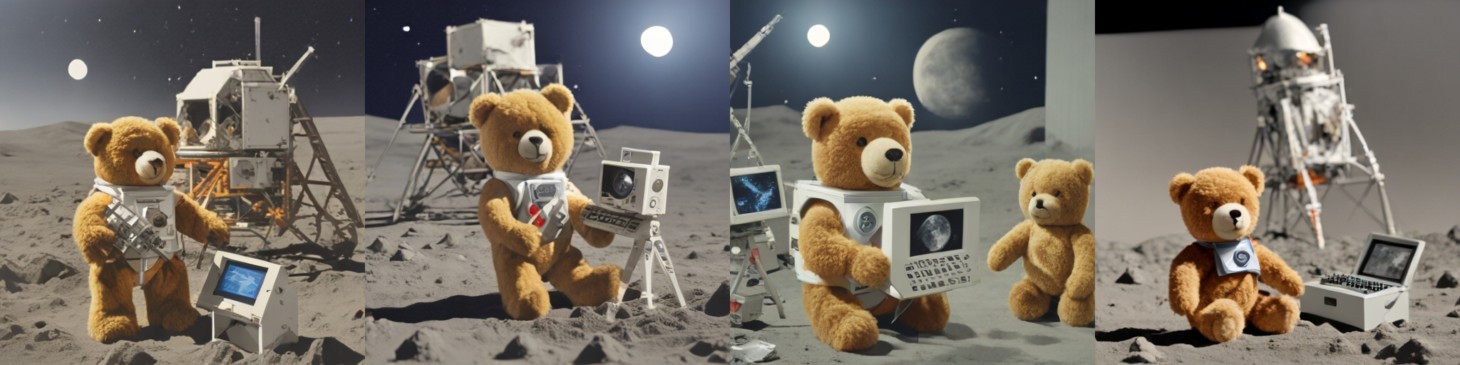}
  \\
  \rotatebox[origin=c]{90}{\makecell{LCM-XL \\ (4 steps)}} &
  \includegraphics[width=.45\linewidth,valign=m]{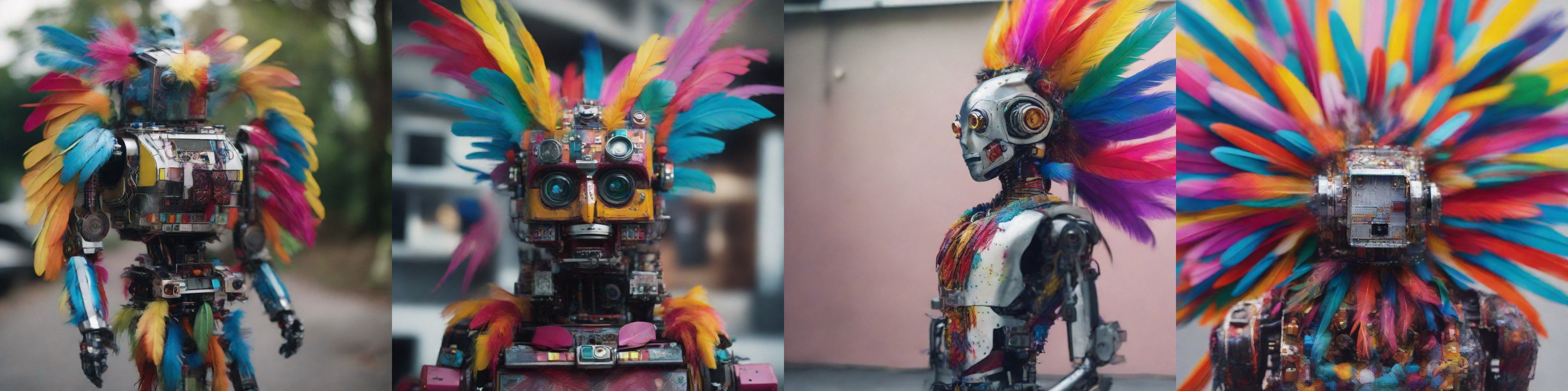}
  &
  \includegraphics[width=.45\linewidth,valign=m]{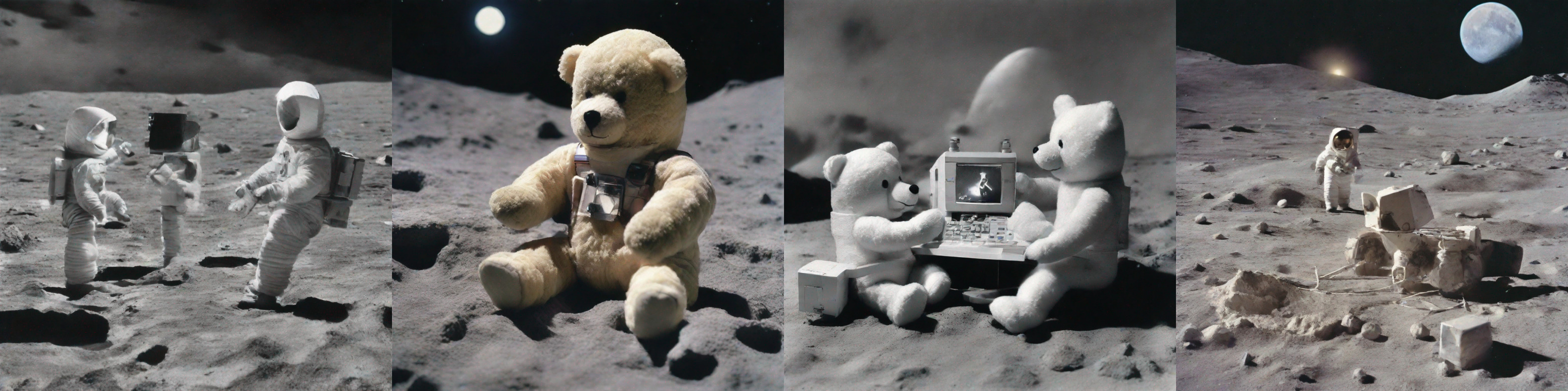}
  \\
  \rotatebox[origin=c]{90}{\makecell{InstaFlow\\ (1 step)}} &
  \includegraphics[width=.45\linewidth,valign=m]{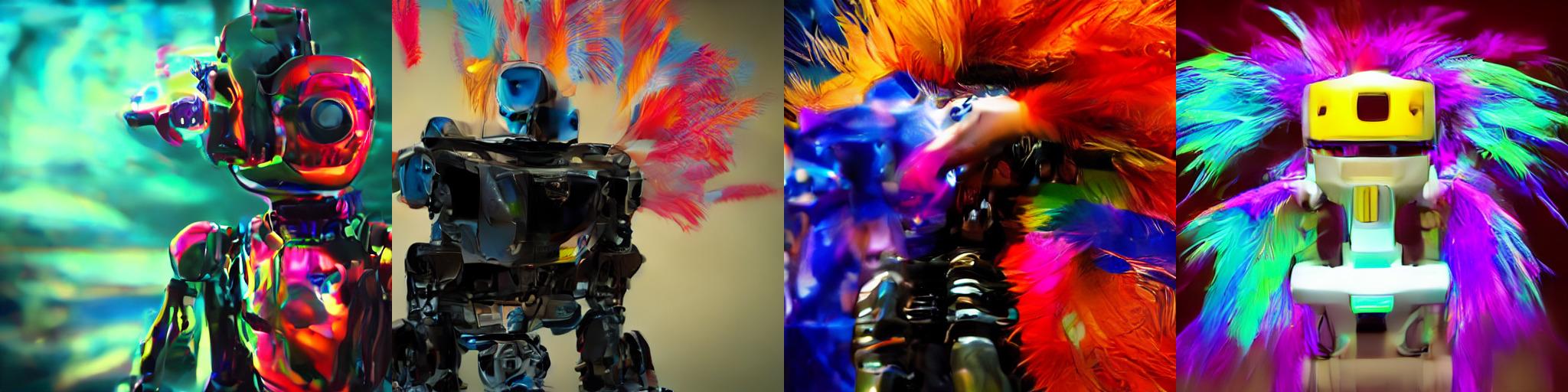}
  &
  \includegraphics[width=.45\linewidth,valign=m]{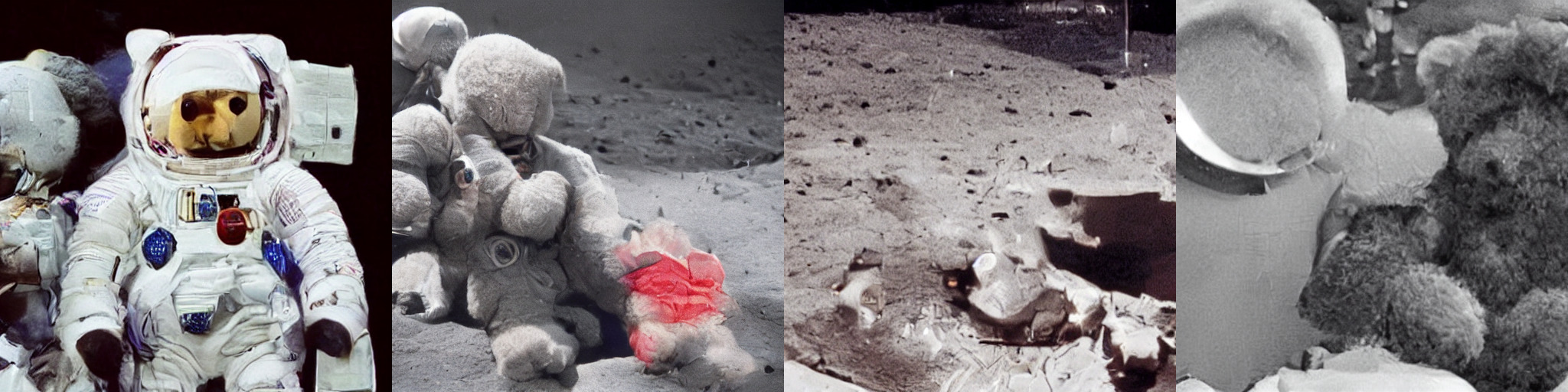}
  \\
    \rotatebox[origin=c]{90}{\makecell{OpenMUSE\\ (16 steps)}} &
  \includegraphics[width=.45\linewidth,valign=m]{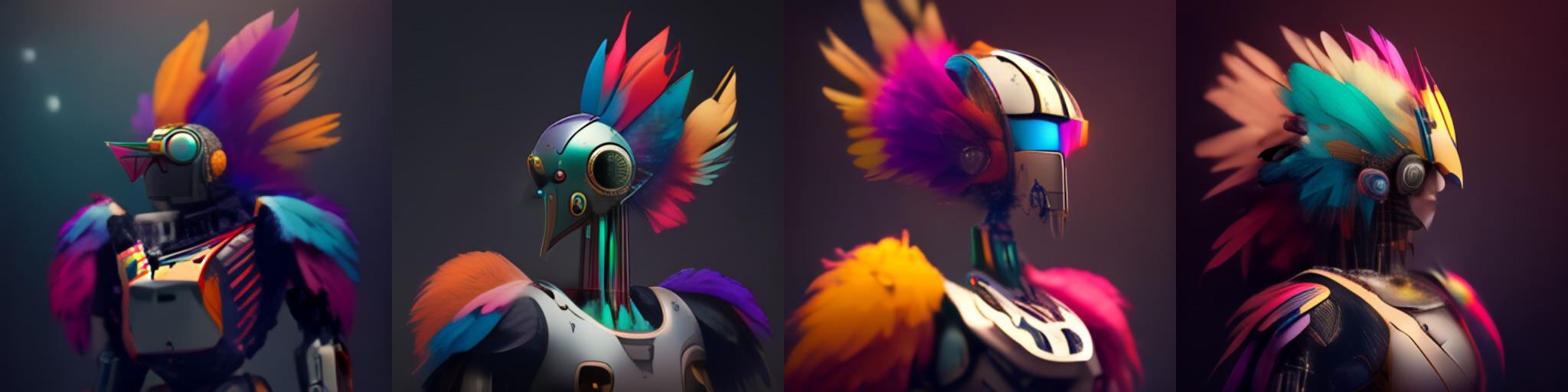}
  &
  \includegraphics[width=.45\linewidth,valign=m]{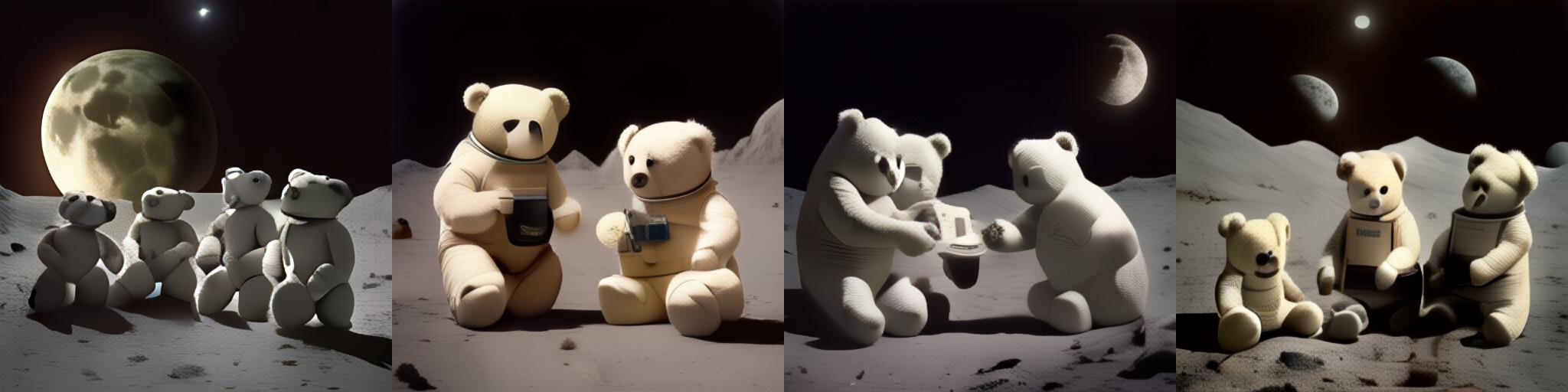}
  \\

\end{tabular}
\caption{
\textbf{Additional qualitative comparisons to state of the art fast samplers.}
Few step samples from our \modelshort-XL and LCM-XL~\citep{Luo2023LCMLoRAAU}, 
InstaFlow~\citep{liu2023instaflow}, and OpenMuse~\cite{patil2023amused}.
\label{fig:qualitativecompsupp}
}
\end{figure*}
}

\newcommand{\s}{\hphantom{0}}

\newcommand{\addablation}{
\begin{table*}[htbp]
\vspace{-.2em}
\centering
\subfloat[
\textbf{Discriminator feature networks}. 
Small, modern DINO networks perform best.
\label{tab:abl0}
]{
\centering
\begin{minipage}{0.29\linewidth}{
\begin{center}
\tablestyle{4pt}{1.05}
\begin{tabular}{y{24}y{28}x{24}x{24}}
Arch & Objective & FID $\downarrow$ & CS $\uparrow$ \\
\shline
ViT-S & DINOv1 & 21.5 & 0.312  \\
ViT-S & DINOv2 &\baseline{\textbf{20.6}} & \baseline{\textbf{0.319}} \\
ViT-L & DINOv2 & 24.0 & 0.302 \\
ViT-L & CLIP   & 23.3 & 0.308 \\
\end{tabular}
\end{center}}\end{minipage}
}
\hspace{2em}
\subfloat[
\textbf{Discriminator conditioning}. Combining image and text conditioning is most effective.
\label{tab:abl1}
]{
\begin{minipage}{0.29\linewidth}{\begin{center}
\tablestyle{4pt}{1.05}
\begin{tabular}{x{24}x{24}x{24}x{24}}
$c_{\mathrm{text}}$  & $c_{\mathrm{img}}$ &  FID $\downarrow$ & CS $\uparrow$ \\
\shline
\xmark & \xmark & 21.2 & 0.302 \\
\cmark & \xmark & 21.2 & 0.307 \\
\xmark & \cmark & 21.1 & 0.316 \\
\cmark & \cmark &\baseline{\textbf{20.6}} & \baseline{\textbf{0.319}} \\
\end{tabular}
\end{center}}\end{minipage}
}
\hspace{2em}
\subfloat[
\textbf{Student pretraining}.
A randomly initialized student network collapses.
\label{tab:abl2}
]{
\begin{minipage}{0.29\linewidth}{\begin{center}
\tablestyle{1pt}{1.05}
\begin{tabular}{y{56}x{24}x{24}}
Initialization & FID $\downarrow$ & CS $\uparrow$ \\
\shline
{Random} & 293.6 & 0.065 \\
{Pretrained}  &\baseline{\textbf{\s20.6}} & \baseline{\textbf{0.319}} \\
\multicolumn{3}{c}{~}\\
\multicolumn{3}{c}{~}\\
\end{tabular}
\end{center}}\end{minipage}
}
\\
\centering
\vspace{.3em}
\subfloat[
\textbf{Loss terms}. 
Both losses are needed and exponential weighting of $\mathcal{L}_{\text{distill}}$ is beneficial. 
\label{tab:abl3}
]{
\begin{minipage}{0.29\linewidth}{\begin{center}
\tablestyle{6pt}{1.05}
\begin{tabular}{y{60}x{24}x{24}}
Loss & FID $\downarrow$ & CS $\uparrow$ \\
\shline
$\mathcal{L}_{\text{adv}}$ & \s20.8 & 0.315 \\
$\mathcal{L}_{\text{distill}}$ & 315.6 & 0.076 \\
$\mathcal{L}_{\text{adv}} + \lambda \mathcal{L}_{\text{distill,exp}}$ & \baseline{\textbf{\s20.6}} & \baseline{0.319} \\
$\mathcal{L}_{\text{adv}} + \lambda \mathcal{L}_{\text{distill,sds}}$ & \s22.3 & 0.325 \\
$\mathcal{L}_{\text{adv}} + \lambda \mathcal{L}_{\text{distill,nfsd}}$ & \s21.8 & \textbf{0.327} \\
\end{tabular}
\end{center}}\end{minipage}
}
\hspace{2em}
\subfloat[
\textbf{Teacher type}.
The student adopts the teacher's traits (SDXL has higher FID \& CS).
\label{tab:abl4}
]{
\centering
\begin{minipage}{0.29\linewidth}{\begin{center}
\tablestyle{4pt}{1.05}
\begin{tabular}{z{28}z{28}x{24}x{24}}
Student & Teacher& FID $\downarrow$ & CS $\uparrow$ \\
\shline
SD2.1 & SD2.1 & \baseline{\textbf{20.6}} & \baseline{0.319} \\
SD2.1 & SDXL  & 21.3 & 0.321 \\
SDXL  & SD2.1 & 29.3 & 0.314 \\
SDXL  & SDXL  & 28.41 & \textbf{0.325} \\
\multicolumn{4}{c}{~}\\
\end{tabular}
\end{center}}\end{minipage}
}
\hspace{2em}
\subfloat[
\textbf{Teacher steps}. A single teacher step is sufficient.
\label{tab:abl5}
]{
\begin{minipage}{0.29\linewidth}{\begin{center}
\tablestyle{4pt}{1.05}
\begin{tabular}{x{18}x{24}x{24}}
Steps & FID $\downarrow$ & CS $\uparrow$ \\
\shline
1 & \baseline{\textbf{20.6}} & \baseline{0.319} \\
2 & 20.8 & \textbf{0.321} \\
4 & 20.3 & 0.317 \\
\multicolumn{3}{c}{~}\\
\multicolumn{3}{c}{~}\\
\end{tabular}
\end{center}}\end{minipage}
}
\vspace{-.1em}
\caption{
\textbf{ADD ablation study.}
We report COCO zero-shot FID\textsubscript{5k} (FID) and CLIP score (CS).
The results are calculated for a single student  step.
The default training settings are:  
DINOv2 ViT-S as the feature network, text and image conditioning for the discriminator, pretrained student weights, a small teacher and student model, and a single teacher step.
The training length is 4000 iterations with a batch size of 128. 
Default settings are marked in \colorbox{baselinecolor}{gray}.
}
\label{tab:ablations} \vspace{-.5em}
\end{table*}
}

\newcommand{\distillcomp}{
\begin{table}[htbp]
\scriptsize
    \centering
    \begin{tabular}{lcccc}
    \toprule
    Method  & \#Steps & Time (s) & FID $\downarrow$ & CLIP $\uparrow$ \\ \midrule
    \multirow{2}{*}{DPM Solver~\cite{lu2022dpm}} & 25 & 0.88  & 20.1  & 0.318  \\ 
     & 8 & 0.34  &  31.7 & 0.320  \\ 
    \midrule
    \multirow{3}{*}{Progressive Distillation~\cite{meng2023distillation}} &1 & 0.09 & 37.2 & 0.275 \\ 
    & 2 & 0.13  & 26.0 & 0.297  \\
    & 4 & 0.21  & 26.4 & 0.300 \\  \midrule
    CFG-Aware Distillation~\cite{li2023snapfusion} & 8& 0.34 & 24.2 & 0.300 \\
    \midrule 
    InstaFlow-0.9B~\cite{liu2023instaflow} & 1 & 0.09 & 23.4  &0.304  \\
    InstaFlow-1.7B~\cite{liu2023instaflow} & 1 & 0.12 & 22.4  & 0.309 \\
    \midrule
    UFOGen~\cite{Xu2023UFOGenYF} & 1 & 0.09  & 22.5  & 0.311 \\
    \midrule
    ADD-M & 1 & 0.09  & \textbf{19.7}  & \textbf{0.326} \\

    \bottomrule
    \end{tabular}
    \caption{\textbf{Distillation Comparison}
    We compare ADD to other distillation methods via 
    COCO zero-shot FID\textsubscript{5k} (FID) and CLIP score (CS). 
    All models are based on SD1.5.
    }
    \label{tab:full-5k}
\end{table}
}
\usepackage{xspace}
\newcommand{\modelshort}{ADD\xspace} 

\newcommand{\figref}[1]{Fig.~\ref{#1}}
\newcommand{\secref}[1]{Section~\ref{#1}}

\newcommand{\eqnref}[1]{Eq.~\eqref{#1}}

\newcommand{\tabref}[1]{Table~\ref{#1}}

\makeatletter
\DeclareRobustCommand\onedot{\futurelet\@let@token\@onedot}
\def\@onedot{\ifx\@let@token.\else.\null\fi\xspace}
\def\eg{e.g\onedot}

\makeatother

\newcommand{\boldparagraph}[1]{\vspace{0.2cm}\noindent{\bf #1} }

\definecolor{darkgreen}{rgb}{0,0.7,0}
\definecolor{darkblue}{RGB}{31,119,180}
\definecolor{darkred}{RGB}{214,39,40}

\usepackage{appendix}

\newlength\savewidth\newcommand\shline{\noalign{\global\savewidth\arrayrulewidth
  \global\arrayrulewidth 1pt}\hline\noalign{\global\arrayrulewidth\savewidth}}
\newcommand{\tablestyle}[2]{\setlength{\tabcolsep}{#1}\renewcommand{\arraystretch}{#2}\centering\footnotesize}
\renewcommand{\paragraph}[1]{\vspace{1.25mm}\noindent\textbf{#1}}

\newcolumntype{x}[1]{>{\centering\arraybackslash}p{#1pt}}
\newcolumntype{y}[1]{>{\raggedright\arraybackslash}p{#1pt}}
\newcolumntype{z}[1]{>{\raggedleft\arraybackslash}p{#1pt}}

\newcommand{\app}{\raise.17ex\hbox{$\scriptstyle\sim$}}

\definecolor{deemph}{gray}{0.6}

\definecolor{baselinecolor}{gray}{.9}
\newcommand{\baseline}[1]{\cellcolor{baselinecolor}{#1}}

\usepackage{pifont}
\newcommand{\cmark}{\ding{51}}%
\newcommand{\xmark}{\ding{55}}%

\definecolor{cvprblue}{rgb}{0.21,0.49,0.74}
\usepackage[pagebackref,breaklinks,colorlinks,citecolor=cvprblue]{hyperref}

\def\paperID{*****} %
\def\confName{CVPR}
\def\confYear{2024}

\title{\vspace{-2em} Adversarial Diffusion Distillation}

\author{
Axel Sauer \and
Dominik Lorenz \and 
Andreas Blattmann \and 
Robin Rombach\\
}

\begin{document}

\twocolumn[{%
\vspace{-2em}
\maketitle%
\vspace{-2em}
\centering
\begin{tabular}{c}
    \large Stability AI \\
    \addlinespace[5pt]
    \small {\emph{Code}: \url{https://github.com/Stability-AI/generative-models} \quad \emph{Model weights}: \url{https://huggingface.co/stabilityai/}}\\
    \addlinespace[5pt]
\end{tabular}
\\
\teaser%
}]

\begin{abstract}
We introduce Adversarial Diffusion Distillation (ADD), a novel training approach that efficiently samples large-scale foundational image diffusion models in just 1--4 steps while maintaining high image quality. 
We use score distillation to leverage large-scale off-the-shelf image diffusion models as a teacher signal in combination with an adversarial loss to ensure high image fidelity even in the low-step regime of one or two sampling steps. 
Our analyses show that our model clearly outperforms existing few-step methods (GANs, Latent Consistency Models) in a single step and reaches the performance of state-of-the-art diffusion models (SDXL) in only four steps.
ADD is the first method to unlock single-step, real-time image synthesis with foundation models.
\end{abstract}    
\section{Introduction}
\label{sec:intro}
Diffusion models (DMs)~\cite{Ho2020DenoisingDP,SohlDickstein2015DeepUL,Song2020ScoreBasedGM} have taken a central role in the field of generative modeling and have recently enabled remarkable advances in 
high-quality image-~\cite{Ramesh2022HierarchicalTI,Rombach2021HighResolutionIS,Balaji2022eDiffITD} and video-~\cite{Ho2022ImagenVH,Blattmann2023AlignYL,esser2023structure} synthesis. One of the key strengths of DMs is their scalability and iterative nature, which allows them to handle complex tasks such as image synthesis from free-form text prompts. 
However, the iterative inference process in DMs requires a significant number of sampling steps, which currently hinders their real-time application. 
Generative Adversarial Networks (GANs)~\cite{Goodfellow2014GenerativeAN, Karras2018ASG,Karras2019AnalyzingAI}, on the other hand, are characterized by their single-step formulation and inherent speed. 
But despite attempts to scale to large datasets\cite{Sauer2022StyleGANXLSS,kang2023scaling}, GANs often fall short of DMs in terms of sample quality. 
The aim of this work is to combine the superior sample quality of DMs with the inherent speed of GANs.

Our approach is conceptually simple: We propose \emph{Adversarial Diffusion Distillation} (\modelshort), a general approach that reduces the number of inference steps of a pre-trained diffusion model to 1--4 sampling steps while maintaining high sampling fidelity and potentially further improving the overall performance of the model. To this end, we introduce a combination of two training objectives: (i) an \textit{adversarial loss} and (ii) a distillation loss that corresponds to \textit{score distillation sampling} (SDS)~\cite{poole2022dreamfusion}.
The adversarial loss forces the model to directly generate samples that lie on the manifold of real images at each forward pass, avoiding blurriness and other artifacts typically observed in other distillation methods~\cite{meng2023distillation}. 
The distillation loss uses another pretrained (and fixed) DM as a teacher to effectively utilize the extensive knowledge of the pretrained DM and preserve the strong compositionality observed in large DMs. %
During inference, our approach does not use classifier-free guidance~\cite{Ho2022ClassifierFreeDG}, further reducing memory requirements. We retain the model's ability to improve results through iterative refinement, which is an advantage over previous one-step GAN-based approaches~\cite{sauer2023stylegan}.

Our contributions can be summarized as follows: 
\begin{itemize}
\item We introduce \modelshort, a method for turning pretrained diffusion models into high-fidelity, real-time image generators using only 1--4 sampling steps. 
\item Our method uses a novel combination of adversarial training and score distillation, for which we carefully ablate several design choices. 
\item \modelshort significantly outperforms strong baselines such as LCM, LCM-XL~\cite{Luo2023LatentCM} and single-step GANs \cite{sauer2023stylegan}, and is able to handle complex image compositions while maintaining high image realism at only a single inference step.
\item Using four sampling steps, \modelshort-XL outperforms its teacher model SDXL-Base at a resolution of $512^2$ px.
\end{itemize}
\section{Background}
\label{sec:background}

While diffusion models achieve remarkable performance in synthesizing and editing high-resolution images~\cite{Ramesh2022HierarchicalTI,Rombach2021HighResolutionIS,Balaji2022eDiffITD} and videos~\cite{Ho2022ImagenVH,Blattmann2023AlignYL}, their iterative nature hinders real-time application.
Latent diffusion models~\cite{Rombach2021HighResolutionIS} attempt to solve this problem by representing images in a more computationally feasible latent space~\cite{Esser2020TamingTF}, but they still rely on the iterative application of large models with billions of parameters.
\system
In addition to utilizing faster samplers for diffusion models ~\cite{song2021denoising, lu2022dpm, dockhorn2022genie, zhang2022fast}, there is a growing body of research on model distillation such as progressive distillation~\cite{salimans2022} and guidance distillation~\cite{meng2023distillation}. These approaches reduce the number of iterative sampling steps to 4-8, but may significantly lower the original performance. Furthermore, they require an iterative training process. 
Consistency models~\cite{Song2023ConsistencyM} address the latter issue by enforcing a consistency regularization on the ODE trajectory and demonstrate strong performance for pixel-based models in the few-shot setting. LCMs~\cite{Luo2023LatentCM} focus on distilling latent diffusion models and achieve impressive performance at 4 sampling steps. Recently, LCM-LoRA~\cite{Luo2023LCMLoRAAU} introduced a low-rank adaptation~\cite{Hu2021LoRALA} training for efficiently learning LCM modules, which can be plugged into different checkpoints for SD and SDXL~\cite{Rombach2021HighResolutionIS,podell2023sdxl}.
InstaFlow \cite{liu2023instaflow} propose to use Rectified Flows \cite{liu2022flow} to facilitate a better distillation process.

All of these methods share common flaws: samples synthesized in four steps often look blurry and exhibit noticeable artifacts. At fewer sampling steps, this problem is further amplified.
GANs~\cite{Goodfellow2014GenerativeAN} can also be trained as standalone single-step models for text-to-image synthesis~\cite{sauer2023stylegan,kang2023scaling}.
Their sampling speed is impressive, yet the performance lags behind diffusion-based models.
In part, this can be attributed to the finely balanced GAN-specific architectures necessary for stable training of the adversarial objective. Scaling these models and integrating advances in neural network architectures without disturbing the balance is notoriously challenging.
Additionally, current state-of-the-art text-to-image GANs do not have a method like classifier-free guidance available which is crucial for DMs at scale. 

\qualitativecomp
Score Distillation Sampling~\cite{poole2022dreamfusion} also known as Score Jacobian Chaining \cite{wang2023score} is a recently proposed method that has been developed to distill the knowledge of foundational T2I Models into 3D synthesis models.
While the majority of SDS-based works~\cite{poole2022dreamfusion, wang2023score, Wang2023ProlificDreamerHA, Metzer2022LatentNeRFFS} use SDS in the context of per-scene optimization for 3D objects, the approach has also been applied to text-to-3D-video-synthesis~\cite{singer2023text} and in the context of image editing~\cite{hertz2023delta}.

Recently, the authors of {\cite{franceschi2023unifying} have shown a strong relationship between score-based models and GANs and propose Score GANs, which are trained using score-based diffusion flows from a DM instead of a discriminator. Similarly, Diff-Instruct \cite{luo2023diff}, a method which generalizes SDS, enables to distill a pretrained diffusion model into a generator without discriminator.

Conversely, there are also approaches which aim to improve the diffusion process using adversarial training.
For faster sampling, Denoising Diffusion GANs \cite{xiao2021tackling} are introduced as a method to enable sampling with few steps.
To improve quality, a discriminator loss is added to the score matching objective in Adversarial Score Matching \cite{jolicoeur2020adversarial} and the consistency objective of CTM \cite{kim2023consistency}.

Our method combines adversarial training and score distillation in a hybrid objective to address the issues in current top performing few-step generative models.

\section{Method}
\label{sec:method}
\figqualitativesteps

Our goal is to generate high-fidelity samples in as few sampling steps as possible, while matching the quality of state-of-the-art models~\citep{podell2023sdxl,Ramesh2022HierarchicalTI,saharia2022photorealistic,dai2023emu}.
The adversarial objective~\citep{Goodfellow2014GenerativeAN,schmidhuber2020generative} naturally lends itself to fast generation as it trains a model that outputs samples on the image manifold in a single forward step.
However, attempts at scaling GANs to large datasets~\cite{sauer2023stylegan,Sauer2022StyleGANXLSS} observed that is critical to not solely rely on the discriminator, 
but also employ a pretrained classifier or CLIP network for improving text alignment.
As remarked in~\cite{sauer2023stylegan}, overly utilizing discriminative networks introduces artifacts and image quality suffers.
Instead, we utilize the gradient of a pretrained diffusion model via a score distillation objective to improve text alignment and sample quality.
Furthermore, instead of training from scratch, we initialize our model with pretrained diffusion model weights; pretraining the generator network is known to significantly improve training with an adversarial loss~\cite{grigoryev2022and}.
Lastly, instead of utilizing a decoder-only architecture used for GAN training~\cite{Karras2018ASG,Karras2019AnalyzingAI}, 
we
adapt
a standard diffusion model framework. This setup naturally enables iterative refinement.

\subsection{Training Procedure}
Our training procedure is outlined in Fig.~\ref{fig:system} and involves three networks:
The \modelshort-student is initialized from a pretrained UNet-DM with weights $\theta$, a discriminator with trainable weights $\phi$, and a DM teacher with frozen weights $\psi$.
During training, the ADD-student generates samples $\hat{x}_{\theta}(x_s, s)$ from noisy data $x_s$.
The noised data points are produced from a dataset of real images $x_0$ via a forward diffusion process $x_s = \alpha_s x_0 +
\sigma_s \epsilon$.  
In our experiments, we use the same coefficients
$\alpha_s$ and $\sigma_s$ as the student DM and sample $s$ uniformly from a set $T_{\mathrm{student}} = \{\tau_1, ..., \tau_n\}$ of $N$ chosen student timesteps. In practice, we choose $N=4$.
Importantly, we set $\tau_n=1000$ and enforce zero-terminal SNR~\citep{lin2023common} during training, as the model needs to start from pure noise during inference.

For the adversarial objective, the generated samples $\hat{x}_{\theta}$ and
real images $x_0$ are passed to the discriminator which aims to distinguish
between them. The design of the discriminator and the adversarial loss are
described in detail in Sec.~\ref{sec:advloss}.
To distill knowledge from the DM teacher, we
diffuse student samples $\hat{x}_{\theta}$ with the teacher's forward process to
$\hat{x}_{\theta, t}$, and use the teacher's denoising prediction
$\hat{x}_{\psi} (\hat{x}_{\theta, t}, t)$ as a reconstruction target for the distillation loss $\mathcal{L}_{\mathrm{distill}}$, see~\secref{sec:distillme}.
Thus, the overall objective is
\begin{equation}
\mathcal{L} =  \mathcal{L}_{\mathrm{adv}}^{\mathrm G}(\hat{x}_{\theta}(x_s,s), \phi) + \lambda \mathcal{L}_{\mathrm{distill}}(\hat{x}_{\theta}(x_s,s), \psi) 
\label{eq:total_loss}
\end{equation}

While we formulate our method in pixel space, it is straightforward to adapt it to LDMs operating in latent space.
When using LDMs with a shared latent space for teacher and student, the distillation loss can be computed in pixel or latent space. 
We compute the distillation loss in pixel space as this yields more stable gradients when distilling latent diffusion model~\cite{yao2023artic3d}.

\humanevalallsingle
\subsection{Adversarial Loss}\label{sec:advloss}
For the discriminator, we follow the proposed design and training procedure in~\cite{sauer2023stylegan} which we briefly summarize; for details, we refer the reader to the original work. 
We use a frozen pretrained feature network $F$ and a set of trainable lightweight discriminator heads $\mathcal{D}_{\phi, k}$. For the feature network $F$, Sauer et al.~\cite{sauer2023stylegan} find vision transformers (ViTs)~\cite{dosovitskiy2020image} to work well, and we ablate different choice for the ViTs objective and model size in \secref{sec:experiments}.
The trainable discriminator heads are applied on features $F_k$ at different layers of the feature network.

To improve performance, the discriminator can be conditioned on additional information via projection~\cite{miyato2018cgans}. 
Commonly, a text embedding $c_{\mathrm{text}}$ is used in the text-to-image setting. But, in contrast to standard GAN training, our training configuration also allows to condition on a given image.
For $\tau < 1000$, the ADD-student receives some signal from the input image $x_0$.
Therefore, for a given generated sample $\hat{x}_{\theta}(x_s, s)$, we can condition the discriminator on information from $x_0$.
This encourages the ADD-student to utilize the input effectively.
In practice, we use an additional feature network to extract an image embedding $c_{\mathrm{img}}$.

Following \cite{sauer2021projected,sauer2023stylegan}, we use the hinge loss~\cite{lim2017geometric} as the adversarial objective function.
Thus the ADD-student's adversarial objective $\mathcal{L}_{\mathrm{adv}}(\hat{x}_{\theta}(x_s,s), \phi)$ amounts to
\begin{equation}
\begin{split}
&\mathcal{L}_{\mathrm{adv}}^{\mathrm G}(\hat{x}_{\theta}(x_s,s), \phi) \\
 &\qquad= - \mathbb{E}_{s, \epsilon, x_0}  \Big[\sum_k \mathcal{D}_{\phi, k} (F_k(\hat{x}_{\theta}(x_s,s)))\Big]\,,
 \end{split}
\end{equation}
whereas the discriminator is trained to minimize
\begin{equation}
\begin{split}
 \mathcal{L}_{\mathrm{adv}}^{\mathrm D}&(\hat{x}_{\theta}(x_s,s), \phi) \\
 = \quad&\mathbb{E}_{x_0} \Big[\sum_k \mathrm{max}(0, 1 - \mathcal{D}_{\phi, k} (F_k(x_0)))  + \gamma R1(\phi)\Big] \\
  +\,&\mathbb{E}_{\hat{x}_\theta} \Big[\sum_k \mathrm{max}(0, 1 + \mathcal{D}_{\phi, k}  (F_k(\hat{x}_\theta))) \Big]\,,
\end{split}
\end{equation}
where $R1$ denotes the R1 gradient penalty~\cite{mescheder2018training}.
Rather than computing the gradient penalty with respect to the pixel values, we compute it on the input of each discriminator head $\mathcal{D}_{\phi, k}$. We find that the $R1$ penalty is particularly beneficial when training at output resolutions larger than $128^2$ px.

\subsection{Score Distillation Loss}
\label{sec:distillme}
\enlargethispage{\baselineskip}
The distillation loss in Eq.~\eqref{eq:total_loss} is formulated as
\begin{equation}
\begin{split}
\mathcal{L}_{\mathrm{distill}}&(\hat{x}_{\theta}(x_s,s), \psi) \\
&=  \mathbb{E}_{t, \epsilon'} \big[c(t) d(\hat{x}_{\theta}, \hat{x}_\psi(\mathrm{sg}(\hat{x}_{\theta, t}); t))\big]\,,
\label{eq:distill_loss}
\end{split}
\end{equation}
where sg denotes the stop-gradient operation.
Intuitively, the loss uses a distance metric $d$ to measure the mismatch between generated samples $x_{\theta}$ by the ADD-student and the DM-teacher's outputs $\hat{x}_{\psi} (\hat{x}_{\theta, t}, t) = (\hat{x}_{\theta, t} - \sigma_t \hat{\epsilon}_\psi(\hat{x}_{\theta, t} , t)) / \alpha_t$ averaged over timesteps $t$ and noise $\epsilon'$.
Notably, the teacher is not directly applied on generations $\hat{x}_{\theta}$ of the ADD-student but instead on diffused outputs $\hat{x}_{\theta, t} =  \alpha_t \hat{x}_{\theta} + \sigma_t \epsilon'$, as non-diffused inputs would be out-of-distribution for the teacher model~\cite{wang2023score}.
\enlargethispage{\baselineskip}%

In the following, we define the distance function $d(x,y) \coloneqq ||x - y||_2^2$. 
Regarding the weighting function $c(t)$, we consider two options: exponential weighting, where $c(t) = \alpha_t$ (higher noise levels contribute less), and score distillation sampling (SDS) weighting~\cite{poole2022dreamfusion}.
In the supplementary material, we demonstrate that with $d(x,y) = ||x - y||_2^2$ and a specific choice for $c(t)$, our distillation loss becomes equivalent to the SDS objective $\mathcal{L}_{\mathrm{SDS}}$, as proposed in \cite{poole2022dreamfusion}. 
The advantage of our formulation is its ability to enable direct visualization of the reconstruction targets and that it naturally facilitates the execution of several consecutive denoising steps.
Lastly, we also evaluate noise-free score distillation (NFSD) objective, a recently proposed  variant of SDS~\cite{katzir2023noise}.    
\section{Experiments}
\label{sec:experiments}
\addablation
For our experiments, we train two models of different capacities, ADD-M (860M parameters) and ADD-XL (3.1B parameters). 
For ablating ADD-M, we use a Stable Diffusion (SD) 2.1 backbone~\cite{Rombach2021HighResolutionIS}, and for fair comparisons with other baselines, we use SD1.5.
ADD-XL utilizes a SDXL~\cite{podell2023sdxl} backbone.
All experiments are conducted at a standardized resolution of 512x512 pixels; outputs from models generating higher resolutions are down-sampled to this size.

We employ a distillation weighting factor of \(\lambda = 2.5\) across all experiments. Additionally, the R1 penalty strength $\gamma$ is set to $10^{-5}$.
For discriminator conditioning, we use a pretrained CLIP-\mbox{ViT-g-14} text encoder~\cite{radford2021learning} to compute text embeddings $c_{\mathrm{text}}$ and the CLS embedding of a DINOv2 ViT-L encoder~\cite{oquab2023dinov2} for image embeddings $c_{\mathrm{img}}$.
For the baselines, we use the best publicly available models:
Latent diffusion models~\cite{Rombach2021HighResolutionIS, podell2023sdxl} (SD1.5\footnote{\url{https://github.com/CompVis/stable-diffusion}}, SDXL\footnote{\url{https://github.com/Stability-AI/generative-models}}) 
cascaded pixel diffusion models~\cite{saharia2022photorealistic} (IF-XL\footnote{\url{https://github.com/deep-floyd/IF}}), distilled diffusion models~\cite{luo2023latent,luo2023lcm} (LCM-1.5, LCM-1.5-XL\footnote{\url{https://huggingface.co/latent-consistency/lcm-lora-sdxl}}), and OpenMUSE \footnote{\url{https://huggingface.co/openMUSE}}~\cite{patil2023amused}, a reimplementation of MUSE~\cite{chang2023muse}, a transformer model specifically developed for fast inference.
Note that we compare to the SDXL-Base-1.0 model without its additional refiner model; this is to ensure a fair comparison.
As there are no public state-of-the-art GAN models, we retrain StyleGAN-T~\cite{sauer2023stylegan} with our improved discriminator. This baseline (StyleGAN-T++) significantly outperforms the previous best GANs in FID and CS, see supplementary.
We quantify sample quality via FID \cite{heusel2017gans} and text alignment via CLIP score \cite{hessel2021clipscore}. For CLIP score, we use \mbox{ViT-g-14} model trained on \mbox{LAION-2B}~\mbox{\cite{schuhmann2022laion}}. Both metrics are evaluated on 5k samples from COCO2017~\citep{lin2015microsoft}.

\subsection{Ablation Study}\label{sec:ablation}
\enlargethispage{\baselineskip}
Our training setup opens up a number of design spaces regarding the adversarial loss, distillation loss, initialization, and loss interplay.
We conduct an ablation study on several choices in \tabref{tab:ablations}; key insights are highlighted below each table. 
We will discuss each experiment in the following.

\boldparagraph{Discriminator feature networks.} 
(\tabref{tab:abl0}). Recent insights by Stein et al.~\cite{stein2023exposing} suggest that ViTs trained with the CLIP~\cite{radford2021learning} or DINO~\cite{caron2021emerging, oquab2023dinov2} objectives are particularly well-suited for evaluating the performance of generative models. 
Similarly, these models also seem effective as discriminator feature networks, with DINOv2 emerging as the best choice.

\boldparagraph{Discriminator conditioning.} 
(\tabref{tab:abl1}).
Similar to prior studies, we observe that text conditioning of the discriminator enhances results. 
Notably, image conditioning outperforms text conditioning, and the combination of both $c_{\mathrm{text}}$ and $c_{\mathrm{img}}$ yields the best results.

\boldparagraph{Student pretraining.} 
(\tabref{tab:abl2}).
Our experiments demonstrate the importance of pretraining the ADD-student. 
Being able to use pretrained generators is a significant advantage over pure GAN approaches.
A problem of GANs is the lack of scalability; both Sauer et al.~\cite{sauer2023stylegan} and Kang et al.~\cite{kang2023scaling} observe a saturation of performance after a certain network capacity is reached. This observation contrasts the generally smooth scaling laws of DMs~\cite{peebles2023scalable}.
However, ADD can effectively leverage larger pretrained DMs (see~\tabref{tab:abl2}) and benefit from stable DM pretraining.

\boldparagraph{Loss terms.} 
(\tabref{tab:abl3}).
We find that both losses are essential. 
The distillation loss on its own is not effective, but when combined with the adversarial loss, there is a noticeable improvement in results.
Different weighting schedules lead to different behaviours, the exponential schedule tends to yield more diverse samples, as indicated by lower FID, SDS and NFSD schedules improve quality and text alignment.
While we use the exponential schedule as the default setting in all other ablations, we opt for the NFSD weighting for training our final model.
Choosing an optimal weighting function presents an opportunity for improvement.
Alternatively, scheduling the distillation weights over training, as explored in the 3D generative modeling literature ~\cite{huang2023dreamtime} could be considered.
\humanevalallmultiple
\distillcomp

\boldparagraph{Teacher type.} 
(\tabref{tab:abl4}).
Interestingly, a bigger student and teacher does not necessarily result in better FID and CS. Rather, the student adopts the teachers characteristics.
SDXL obtains generally higher FID, possibly because of its less diverse output, yet it exhibits higher image quality and text alignment~\cite{podell2023sdxl}.

\boldparagraph{Teacher steps.} 
(\tabref{tab:abl5}).
While our distillation loss formulation allows taking several consecutive steps with the teacher by construction, we find that several steps do not conclusively result in better performance.

\eloplot 

\subsection{Quantitative Comparison to State-of-the-Art}
\label{sec:sotacomparison}

For our main comparison with other approaches, we refrain from using automated metrics, as user preference studies are more reliable~\cite{podell2023sdxl}. 
In the study, we aim to assess both prompt adherence and the overall image.
As a performance measure, we compute win percentages for pairwise comparisons and ELO scores when comparing several approaches. 
For the reported ELO scores we calculate the mean scores between both prompt following and image quality.
Details on the ELO score computation and the study parameters are listed in the supplementary material. 

\figref{fig:humanevalsingle} and \figref{fig:humanevalmultiple} present the study results.
The most important results are:
First, ADD-XL outperforms LCM-XL (4 steps) with a single step.
Second, ADD-XL can beat SDXL (50 steps) with four steps in the majority of comparisons.
This makes ADD-XL the state-of-the-art in both the single and the multiple steps setting.
\figref{fig:eloplot} visualizes ELO scores relative to inference speed. Lastly, \tabref{tab:full-5k} compares different few-step sampling and distillation methods using the same base model. ADD outperforms all other approaches including the standard DPM solver with eight steps.

\subsection{Qualitative Results}
\qualitativeteacher
To complement our quantitative studies above, we present qualitative results in this section. To paint a more complete picture, we provide additional samples and qualitative comparisons in the supplementary material.
\figref{fig:qualitativecomp} compares ADD-XL (1 step) against the best current baselines in the few-steps regime. 
\figref{fig:qualitativesteps} illustrates the iterative sampling process of ADD-XL. These results showcase our model's ability to improve upon an initial sample.
Such iterative improvement represents another significant benefit over pure GAN approaches like StyleGAN-T++.
Lastly, \figref{fig:qualitativeteacher} compares ADD-XL directly with its teacher model SDXL-Base. As indicated by the user studies in \secref{sec:sotacomparison}, ADD-XL outperforms its teacher in both quality and prompt alignment. The enhanced realism comes at the cost of slightly decreased sample diversity.

\section{Discussion}
\label{sec:discussion}
\enlargethispage{\baselineskip}
This work introduces \emph{Adversarial Diffusion Distillation}, a general method for distilling a pretrained diffusion model into a fast, few-step image generation model.
We combine an adversarial and a score distillation objective to distill the public Stable Diffusion~\cite{Rombach2021HighResolutionIS} and SDXL~\cite{podell2023sdxl} models, leveraging both real data through the discriminator and structural understanding through the diffusion teacher.
Our approach performs particularly well in the ultra-fast sampling regime of one or two steps, and our analyses demonstrate that it outperforms all concurrent methods in this regime. 
Furthermore, we retain the ability to refine samples using multiple steps. In fact, using four sampling steps, our model outperforms widely used multi-step generators such as SDXL, IF, and OpenMUSE.

Our model enables the generation of high quality images in a single-step, opening up new possibilities for real-time generation with foundation models.

\section*{Acknowledgements}

We would like to thank 
Jonas M\"uller for feedback on the draft, the proof, and typesetting;
Patrick Esser for feedback on the proof and building an early model demo;
Frederic Boesel for generating data and helpful discussions;
Minguk Kang and Taesung Park for providing GigaGAN samples;
Richard Vencu, Harry Saini, and Sami Kama for maintaining the compute infrastructure; 
Yara Wald for creative sampling support; and Vanessa Sauer for her general support. 

\FloatBarrier
{
\small
\bibliographystyle{ieeenat_fullname}
\bibliography{ms}
}

\clearpage
\onecolumn
\appendix

\section*{Appendix}

\section{SDS As a Special Case of the Distillation Loss}
\label{app:sdsproof}
If we set the weighting function to $c(t) = \frac{\alpha_t}{2 \sigma_t} w(t)$ where $w(t)$ is the scaling factor from the weighted diffusion loss as in \cite{poole2022dreamfusion} and choose $d(x,y) = ||x - y||_2^2$, the distillation loss in~\eqnref{eq:distill_loss} is equivalent to the score distillation objective:

\begin{equation}
  \begin{split}
  \frac{d}{d\theta} & \mathcal{L}_{\text{distill}}^{\mathrm{MSE}} \\
    &=  \mathbb{E}_{t, \epsilon'} \Big[ c(t) \frac{d}{d\theta}  \vert \vert \hat{x}_{\theta} - \hat{x}_{\psi} (\text{sg}(\hat{x}_{\theta, t}) ; t)\vert \vert_2 ^2 \Big] \\
    &=  \mathbb{E}_{t, \epsilon'} \Big[ 2 c(t)  [ \hat{x}_{\theta}  - \hat{x}_{\psi} (\hat{x}_{\theta, t} ; t )]\frac{d\hat{x}_{\theta} }{d\theta} \Big] \\
    &=  \mathbb{E}_{t, \epsilon'} \Big[ \frac{\alpha_t}{\sigma_t} w(t)  [\frac{1}{\alpha_t} (\hat{x}_{\theta, t}  - \hat{x}_{\theta, t}) + \hat{x}_{\theta}  - \hat{x}_{\psi} (\hat{x}_{\theta, t} ; t )] \frac{d\hat{x}_{\theta} }{d\theta} \Big] \\
    &=  \mathbb{E}_{t, \epsilon'} \Big[ \frac{1}{\sigma_t} w(t)  [( \alpha_t  \hat{x}_{\theta} - \hat{x}_{\theta, t})  - (\alpha_t\hat{x}_{\psi} (\hat{x}_{\theta, t} ; t) -  \hat{x}_{\theta, t}) ]\frac{d\hat{x}_{\theta} }{d\theta}\Big]  \\
    &=  \mathbb{E}_{t, \epsilon'}  \Big[ \frac{w(t)}{\sigma_t}   [-\sigma_t  \epsilon'  + \sigma_t \hat{\epsilon}_{\theta}(\hat{x}_{\theta, t} ; t)] \frac{d\hat{x}_{\theta} }{d\theta} \Big]  \\
    &= \frac{d}{d\theta} \mathcal{L}_{\text{SDS}}
  \end{split}
\end{equation}
\vspace{2em}

\section{Details on Human Preference Assessment}
\label{app:human_eval}

For the evaluation results presented in \Cref{fig:humanevalsingle,fig:humanevalmultiple,fig:eloplot}, we employ human evaluation and do not rely on commonly used metrics for quality assessment of generative models such as FID~\cite{heusel2017gans} and CLIP-score~\citep{radford2021learning}, since these have been shown to capture more fine grained aspects like aesthetics and scene composition only insufficiently~\citep{podell2023sdxl,kirstain2023pickapic}. 
However these categories in particular have become more and more important when comparing current state-of-the-art text-to-image models. We evaluate all models based on 100 selected prompts from the PartiPrompts benchmark~\citep{yu2022scaling} with the most relevant categories (excluding prompts from the category \emph{basic}). More details on how the study was conducted~\Cref{app:human_eval_setup} and the rankings computed~\Cref{app:elo_score} are listed below.

\humanevalallsingleADDM
\humanevalallmultipleADDM

\subsection{Experimental Setup}
\label{app:human_eval_setup}

Given all models for one particular study (\eg ADD-XL, OpenMUSE\footnote{\url{https://huggingface.co/openMUSE}}, IF-XL\footnote{\url{https://github.com/deep-floyd/IF}}, SDXL~\citep{podell2023sdxl} and LCM-XL\footnote{\url{https://huggingface.co/latent-consistency/lcm-lora-sdxl}}~\citep{Luo2023LatentCM,Luo2023LCMLoRAAU} in \Cref{fig:eloplot}) we compare each prompt for each pair of models (1v1). 
For every comparison, we collect an average of four votes per task from different annotators, for both visual quality and prompt following. Human evaluators, recruited from the platform \emph{Prolific}\footnote{\url{https://app.prolific.com}} with English as their first language, are shown two images from different models based on the same text prompt. To prevent biases, evaluators are restricted from participating in more than one of our studies. For the prompt following task, we display the text prompt above the two images and ask, \enquote{Which image looks more representative of the text shown above and faithfully follows it?} For the visual quality assessment, we do not show the prompt and instead ask, \enquote{Which image is of higher quality and aesthetically more pleasing?}.
Performing a complete assessment between all pair-wise comparisons gives us robust and reliable signals on model performance trends and the effect of varying thresholds. The order of prompts and the order between models are fully randomized. Frequent attention checks are in place to ensure data quality.

\subsection{ELO Score Calculation}
\label{app:elo_score}
To calculate rankings when comparing more than two models based on 1v1 comparisons we use ELO Scores (higher-is-better)~\citep{elo1978rating} which were originally proposed as a scoring method for chess players but have more recently also been applied to compare instruction-tuned generative LLMs~\cite{bai2022training,askell2021general}. For a set of competing players with initial ratings $R_{\text{init}}$ participating in a series of zero-sum games the ELO rating system updates the ratings of the two players involved in a particular game based on the expected and and actual outcome of that game. Before the game with two players with ratings $R_1$ and $R_2$, the expected outcome for the two players are calculated as
\begin{align}
\label{eq:expected_elo}
E_1 =  \frac{1}{1 + 10^{\frac{R_2 - R_1}{400}}} \, , \\
E_2 =  \frac{1}{1 + 10^{\frac{R_1 - R_2}{400}}} \, . 
\end{align}
After observing the result of the game, the ratings $R_i$ are updated via the rule
\begin{align}
\label{eq:ranking_update}
R^{'}_{i} =  R_i + K \cdot \left(S_i - E_i \right), \quad i \in \{1,2\}
\end{align}
where $S_i$ indicates the outcome of the match for player $i$. In our case we have $S_i=1$ if player $i$ wins and $S_i = 0$ if player $i$ looses. The constant $K$ can be see as weight putting emphasis on more recent games. We choose $K=1$ and bootstrap the final ELO ranking for a given series of comparisons based on 1000 individual ELO ranking calculations with randomly shuffled order. Before comparing the models we choose the start rating for every model as $R_{\text{init}} = 1000$.

\section{GAN Baselines Comparison}
\label{app:gan_baselines}
For training our state-of-the-art GAN baseline StyleGAN-T++, we follow the training procedure outlined in~\cite{sauer2023stylegan}. The main differences are extended training ($\sim$2M iterations with a batch size of 2048, which is comparable to GigaGAN's schedule~\cite{kang2023scaling}), the improved discriminator architecture proposed in~\secref{sec:advloss}, and R1 penalty applied at each discriminator head.

\figref{fig:truncation} shows that StyleGAN-T++ outperforms the previous best GANs by achieving a comparable zero-shot FID to GigaGAN at a significantly higher  CLIP score.
Here, we do not compare to DMs, as comparisons between model classes via automatic metrics tend to be less informative~\cite{stein2023exposing}. 
As an example, GigaGAN achieves FID and CLIP scores comparable to SD1.5, but its sample quality is still inferior, as noted by the authors.

\truncation
\suppteaser
\clearpage
\section{Additional Samples}
\label{sec:additionalsamples}

We show additional one-step samples as in \Cref{fig:teaser} in \Cref{fig:suppteaser}. 
An additional qualitative comparison as in \Cref{fig:qualitativesteps} which demonstrates that our model can further refine quality by using more than one sampling step is provided in \Cref{fig:suppsteps}, where we show that, while sampling quality with a single step is already high, more steps can give higher diversity and better spelling capabilities.  
Lastly, we provide an additional qualitative comparison of \modelshort-XL to other state-of-the-art one and few-step models in \Cref{fig:qualitativecompsupp}. 
\qualitativecompsupp
\suppsteps

\end{document}